\documentclass[letterpaper]{article} 
\usepackage{aaai24}  

\usepackage{subcaption}
\usepackage{multirow}
\usepackage{booktabs} 
\usepackage{makecell}
\usepackage{xcolor,colortbl}
\usepackage{nicematrix}
\usepackage{enumitem}
\usepackage{amsmath}
\usepackage{listings}

\usepackage{amsfonts}       
\usepackage{nicefrac}       
\usepackage{microtype}      
\usepackage{xcolor}         

\usepackage{caption}
\usepackage{enumitem}

\usepackage{amsmath,amsfonts,bm}

\def\eqref#1{equation~\ref{#1}}

\def\1{\bm{1}}

\def\rvs{{\mathbf{s}}}

\def\rvy{{\mathbf{y}}}

\DeclareMathAlphabet{\mathsfit}{\encodingdefault}{\sfdefault}{m}{sl}
\SetMathAlphabet{\mathsfit}{bold}{\encodingdefault}{\sfdefault}{bx}{n}

\DeclareMathOperator*{\argmax}{arg\,max}

\newcommand{\keyfact}{informativeness}

\newcommand{\awren}{\ensuremath{\overline{\text{WReN}}}}
\newcommand{\atrans}{\ensuremath{\overline{\text{Trans.}}}}
\newcommand{\mwren}{\ensuremath{\text{WReN}^\star}}
\newcommand{\mtrans}{\ensuremath{\text{Trans.}^\star}}

\usepackage{times}  
\usepackage{helvet}  
\usepackage{courier}  
\usepackage[hyphens]{url}  
\usepackage{graphicx} 
\usepackage{natbib}  
\usepackage{caption} 
\usepackage{algorithm}
\usepackage{algorithmic}

\usepackage{newfloat}
\usepackage{listings}
\DeclareCaptionStyle{ruled}{labelfont=normalfont,labelsep=colon,strut=off} 
\floatstyle{ruled}
\newfloat{listing}{tb}{lst}{}
\floatname{listing}{Listing}
\title{Revisiting Disentanglement in Downstream Tasks: \\ A Study on Its Necessity for Abstract Visual Reasoning}
\author{
    Ruiqian Nai\textsuperscript{\rm 1, \rm 2, \rm 3},
    Zixin Wen\textsuperscript{\rm 4},
    Ji Li\textsuperscript{\rm 1},
    Yuanzhi Li\textsuperscript{\rm 4},
    Yang Gao\textsuperscript{\rm 1, \rm 2, \rm 3}\thanks{Corresponding author.}
}
\affiliations{
    \textsuperscript{\rm 1}Tsinghua University, Beijing, China \\
    \textsuperscript{\rm 2}Shanghai Artificial Intelligence Laboratory, Shanghai, China\\
    \textsuperscript{\rm 3}Shanghai Qi Zhi Institute, Shanghai, China \\
    \textsuperscript{\rm 4} Carnegie Mellon University, Pittsburgh, PA, USA

    \{nrq22, ji-li20\}@mails.tsinghua.edu.cn, \{gaoyangiiis
\}@tsinghua.edu.cn, \{zixinw, yuanzhil\}@andrew.cmu.edu
}

\usepackage{bibentry}

\begin{document}

\maketitle

\begin{abstract}
     In representation learning, a disentangled representation is highly desirable as it encodes generative factors of data in a separable and compact pattern. Researchers have advocated leveraging disentangled representations to complete downstream tasks with encouraging empirical evidence.  This paper further investigates the necessity of disentangled representation in downstream applications. Specifically,  we show that dimension-wise disentangled representations are unnecessary on a fundamental downstream task, abstract visual reasoning. We provide extensive empirical evidence against the necessity of disentanglement, covering multiple datasets, representation learning methods, and downstream network architectures. Furthermore, our findings suggest that the informativeness of representations is a better indicator of downstream performance than disentanglement.  Finally, the positive correlation between informativeness and disentanglement explains the claimed usefulness of disentangled representations in previous works. The source code is available at https://github.com/Richard-coder-Nai/disentanglement-lib-necessity.git.
\end{abstract}
\section{Introduction}
Disentanglement has been considered an essential property of representation learning \citep{bengio2013representation, peters2017elements, goodfellow2016deep, bengio2007scaling, schmidhuber1992learning,lake2017building, tschannen2018recent}. Disentanglement is defined as a dimension-wise relationship, wherein a representation dimension should capture information from exactly one factor and vice versa \citep{locatello2019challenging,higgins2016beta, kim2018disentangling, chen2018isolating, eastwood2018framework, ridgeway2018learning, kumar2017variational, do2019theory}. Such property is analogous to biological mechanisms, as neurons in brains are specialized for specific tasks, with some aligned to axes of data generative factors \citep{higgins2021unsupervised, whittington2022disentangling}. Moreover, disentangled representations offer a compact and separable structure. Therefore, they are believed to aid in compositional generalization and reasoning, potentially leading to improved performance in downstream tasks \citep{bengio2013representation}. 

These purported advantages have been verified on multiple downstream tasks. For instance, abstract visual reasoning \cite{van2019disentangled}, fairness \cite{locatello2019fairness}, and out-of-distribution (OOD) generalization \cite{dittadi2020transfer}. Disentangled representations result in better downstream performance, faster learning, and a strong correlation with task success. Based on these promising empirical results, adopting disentangled representations is considered a wise decision for completing downstream tasks.

However, \citet{cao2022empirical} demonstrates that contrastive learning, does not produce fully disentangled representations.  Yet contrastive pre-training methods achieve remarkable downstream performances \citep{he2020momentum,caron2021emerging,he2022masked,oquab2023dinov2}. These findings may contradict earlier claims regarding the benefits of developing disentangled representations, as such disentanglement does not appear to play a significant role in the success of downstream tasks. This potential conflict with the previously claimed usefulness of disentanglement motivates us to reassess the role of disentanglement for downstream tasks. Are disentangled representations \textit{necessary} for downstream tasks? If not, how do we explain the previously reported benefits?

We choose abstract visual reasoning as the testbed to investigate the necessity of disentanglement.
In this task, intelligent agents are asked to take human IQ tests, i.e., predict the missing panel of  Raven's Progressive Matrices (RPMs) \citep{raven1941standardization}. To solve this task, it is essential to comprehend and employ the generative factors of data, as they serve as reasoning attributes.
The task is completed in a \textbf{two-staged} fashion: \textbf{(1)} extracting RPMs' representations in an unsupervised manner, \textbf{(2)} then performing abstract reasoning training based on learned representations. Therefore, it is an essential and popular benchmark for disentangled representation learning \cite{van2019disentangled,locatello2020weakly,scholkopf2021toward}.  
In practice, among various researched downstream tasks \citep{van2019disentangled, locatello2019fairness, dittadi2020transfer}, disentanglement has been shown to improve sample efficiency and final performance on abstract reasoning tasks \citep{van2019disentangled}. Researchers have therefore suggested the use of disentangled representations \cite{steenbrugge2018improving, van2019disentangled, malkinski2022deep}.

We conduct an extensive empirical study that uses abstract reasoning tasks to investigate the role of disentanglement for downstream tasks. We train 720 representation learning models on two datasets, including disentanglement and general-purpose methods. We then train 5 WReNs \citep{barrett2018measuring} and 5 Transformers \citep{vaswani2017attention,hahne2019attention}  using the outputs of each representation learning model to perform abstract reasoning, yielding a total of 7200 abstract reasoning models. Our contributions are as follows.
\begin{itemize}
    \item We conduct a comprehensive exploration into the impact of disentanglement on downstream tasks, by introducing both disentangled and general-purpose representations and employing multiple methods (WReN and Transformer) to complete the downstream task.
    \item  We show the unnecessity by highlighting the significance of \textit{informativeness} over disentanglement. Informativeness measures \textit{what} information the representation has learned \cite{eastwood2018framework}. Previous studies' analysis on informativeness is scarce and thus overlook its importance \cite{van2019disentangled,locatello2019fairness,dittadi2020transfer,trauble2021role}.
    \item We show that informativeness is the underlying factor behind the previously argued usefulness of disentanglement, as we observe limited extra benefits of disentanglement when informativeness is closely matched.
\end{itemize}

\section{Related Work}
\label{sec:main_related_work}
\textbf{Disentangled representation learning.}  In this paper, we adopt the widely accepted definition of disentanglement: a one-to-one mapping between representation dimensions and generative factors of data, which we term ``dimension-wise disentanglement''. It requires that each representation dimension encode only one factor and vice versa \citep{locatello2019challenging,eastwood2018framework, kumar2017variational, do2019theory}. While some studies suggest relaxing this constraint, introducing new properties, or expanding the definition of disentanglement, \citep{higgins2018towards, wang2021desiderata, roth2022disentanglement, eastwood2022dci}, our focus remains on this well-established definition. 
Another line of work related to disentangled representation learning is the Independent Component Analysis (ICA) \citep{comon1994independent}. ICA aims to recover independent components of the data. 

Based on the dimension-wise definition, researchers develop methods and metrics. SOTA disentanglement methods are mainly variants of generative methods ~\citep{higgins2016beta,kim2018disentangling,burgess2018understanding,kumar2017variational,chen2018isolating, chen2016infogan,jeon2018ib,lin2020infogan,leeb2020structure}.
Corresponding metrics are designed \citep{higgins2016beta,kim2018disentangling,chen2018isolating, eastwood2018framework,kumar2017variational,cao2022empirical}. 


\textbf{Downstream tasks.}  Several works conduct empirical studies on downstream tasks to support the believed benefits of disentanglement \citep{bengio2013representation,do2019theory}, including abstract reasoning \citep{van2019disentangled}, fairness \citep{locatello2019fairness}, and OOD generalization \citep{dittadi2020transfer}. Provided with positive empirical results, these works advocate using disentanglement to complete downstream tasks. 
Among these works, \citet{van2019disentangled} reported the most encouraging evidence from abstract reasoning tasks. Disentanglement is more significant than other representation properties, especially in limited samples. Therefore, we adopt their settings and investigate the necessity of disentanglement on the same tasks at a few-sample regime. However, there are a few issues with their study.
Firstly, it underestimates factors' linear classification accuracy, yielding a weaker correlation between informativeness and downstream performance (see Figure 8 in Appendix B.1).   Moreover, it paid insufficient attention to the analysis of informativeness. We address these issues and show the unnecessity of disentanglement on a broader range of representations and downstream methods. 

\citet{trauble2021role} delves into the role of sim2real transfer in reinforcement learning tasks using real robots, and their findings closely align with ours. However, their scope is still confined to variants of VAEs. Our work is complementary to theirs on a more fundamental and straightforward mechanism. Instead of the sophisticated OOD generalization tasks, we focus on the essential setting of abstract visual reasoning.  

\textbf{Abstract visual reasoning} has been a popular benchmark to measure the representation's downstream performance, especially in disentanglement literature \citep{steenbrugge2018improving,van2019disentangled,dittadi2020transfer,locatello2020weakly,scholkopf2021toward}.  The most common type is the Raven's Progressive Matrices (RPMs) \citep{raven1941standardization}.  To solve RPMs, one is asked to complete the missing panel of a $3\times 3$ grid by exploring the logical relationships of 8 context panels. Moreover, abstract visual reasoning is a well-developed benchmark for representation learning. Given that it is coupled with a principle treatment of generalization \citep{fleuret2011comparing}, a neural network can not solve reasoning tasks by simply memorizing superficial statistical features. 
\section{Downstream Benchmark: Abstract Visual Reasoning}
In this section, we'll introduce the abstract visual reasoning task and outline the downstream benchmark framework, including representation learning methods, metrics, and abstract reasoning models.
\begin{figure}[ht]
  \begin{center}
    \includegraphics[width=0.3\textwidth]{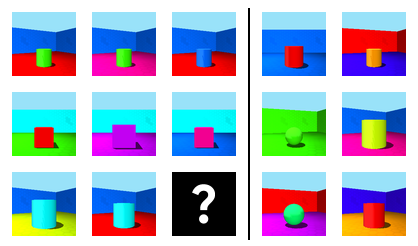}
  \end{center}
  \vspace{-0.5em}
  \caption{An example of RPM on \textit{3DShapes} from \citet{van2019disentangled}. }
  \vspace{-1.5em}
  \label{fig:3dshapes_question}
\end{figure}
\subsection{Abstract Visual Reasoning as a Two-staged Task}
\label{sec:background_abstract_reasoning}
The abstract visual reasoning tasks are highly inspired by the famous human IQ test, Raven's Progressive Matrices (RPMs) \citep{raven1941standardization}. 
Figure~\ref{fig:3dshapes_question} shows an RPM question in our evaluation dataset. 
There are eight context panels and one missing panel in the left part of the figure. The context panels are arranged following some logical rules across rows. During the test, the subject must pick one of  the six candidates in the right part to fix the missing panel. The goal is to maintain the logical relationships given by the contexts. 
More details of RPMs are available in Appendix A.4.

We adopt RPMs as a downstream benchmark following \citet{van2019disentangled}. To measure the necessity of disentanglement for downstream tasks, we separate the evaluation process into two stages: \textbf{(1)} In Stage-1,  representation learning models extract representations from images of which RPMs consist, and \textbf{(2)} in Stage-2, abstract reasoning models predict the missing panels from the frozen representations of contexts and answer candidates. Correspondingly, we denote representation learning models as \textbf{Stage-1 models} while abstract reasoning models as \textbf{Stage-2 models}. 
For Stage-1, we measure the representation properties, including disentanglement and informativeness. A diverse set of Stage-1 and Stage-2 models are trained, yielding multiple samples from the joint distribution of representation metric scores and downstream accuracy. Finally, we study the relationships between representation qualities and downstream performance. We aim to investigate whether more disentangled representations perform better on abstract reasoning tasks.

\subsection{Background of Representation Learning}
\label{sec:representation_learning}
\textbf{Disentangled representation learning methods.} The seminal works of \citet{higgins2016beta} and \citet{chen2016infogan} embody disentanglement by augmenting deep generative models~\citep{kingma2013auto,goodfellow2014generative}. For disentangled representation learning methods, following \citet{van2019disentangled}, we focus on a family of VAEs with disentanglement inductive bias denoted as DisVAEs. DisVAEs' objective summarizes augmentations of SOTA methods. Namely, $\beta$-VAE~\citep{higgins2016beta}, AnnealedVAE~\citep{burgess2018understanding}, $\beta$-TCVAE~\citep{chen2018isolating}, FactorVAE~\citep{kim2018disentangling}, and DIP-VAE~\citep{kumar2017variational}. They achieve disentanglement mainly by encouraging independence between representation dimensions. Please refer to Appendix A.2 for details.

\textbf{General-purpose representation learning methods.} In our study, methods not (explicitly) encouraging disentanglement are called general-purpose methods.  We take a set of BYOL~\citep{grill2020bootstrap} with the same size as DisVAEs as representatives. BYOL is a negative-free contrastive learning method. It creates different ``views" of an image by data augmentation and pulls together their distance in representation space. To avoid collapsing to trivial representations, a predictor appending to one of the siamese encoders and exponential moving average update strategy~\citep{he2020momentum} are employed. It does not encourage disentanglement due to the lack of regularizers. Indeed, the empirical evidence in \citet{cao2022empirical} demonstrates that representations learned by BYOL have weak disentanglement properties.

\textbf{Representation property metrics. }Considered properties of representations cover two axes of metrics: disentanglement metrics and informativeness metrics \citep{eastwood2018framework, eastwood2022dci}. We include \textit{BetaVAE} score~\citep{higgins2016beta},  \textit{FactorVAE} score  ~\citep{kim2018disentangling},  \textit{Mutual Information Gap} ~\citep{chen2018isolating} , \textit{SAP} ~\citep{kumar2017variational}, and \textit{DCI Disentanglement} ~\citep{eastwood2018framework}. \citet{locatello2019challenging} proves their agreement on VAE methods with extensive experiments.  Though their measurements are different, their results are positively correlated. To facilitate comparison across different representation sizes, we also include the \textit{MED} metric \citep{cao2022empirical}. 
On the other hand, informativeness requires representations to encode enough information about factors \cite{eastwood2018framework}. We employ \textit{Logistic Regression} (LR) and \textit{Linear Regression} in this work. They use a linear model to classify or regress the values of generative factors. Given the weak capacity of linear models, a higher LR accuracy or lower regression error ensures that sufficient information is explicitly encoded. However, it does not emphasize a dimension-wise encoding pattern like disentanglement. To distinguish, we term the property indicated by LR and linear regression as \textit{\keyfact{}} \cite{eastwood2018framework}.

\subsection{Background of Methods for Abstract Reasoning}
\label{sec:background_reasoning_methods}
In Stage-1, we extract representations of eight context panels (the left part of Figure~\ref{fig:3dshapes_question}) and six answer candidates (the right part of Figure~\ref{fig:3dshapes_question}). Then in Stage-2,  downstream models perform abstract reasoning from the (frozen) representations. Abstract reasoning models evaluate whether filling the blank panel by a candidate follows the logical rules given by contexts. For a trial $T_i$ of one candidate $a_i \in A = \{a_1, ..., a_6\}$ and eight context panels $C = \{c_1, ..., c_8\}$, its score is calculated as follows: 
\begin{align*}
    Y_i &= \operatorname{Stage2}(\operatorname{Stage1}(T_i)) ,\\
    \operatorname{Stage1}(T_i) &= \{ \operatorname{Stage1}(c_1), \dots, \operatorname{Stage1} (c_8),  \operatorname{Stage1} (a_i) \},
\end{align*}
where $Y_i$ is the score of trial $T_i$, $\operatorname{Stage1}(\cdot), \operatorname{Stage2}(\cdot)$ denote the forward process of the Stage-1 and Stage-2 models, and $\operatorname{Stage1}(T_i)$ is the representations of contexts and candidate $a_i$. After evaluating all trials $\{T_1, T_2, \dots, T_6\}$, the output answer  is $\argmax_i{Y_i}$.

We implement two different structures of Stage-2 models, namely, WReN~\citep{barrett2018measuring} and Transformer~\citep{vaswani2017attention, hahne2019attention}. First, we employ an MLP  or a Transformer to embed an RPM trial. Then, an MLP head predicts a scalar score from the embeddings.

\section{Experiments}
In this section, we conduct a systematic empirical study about representation properties' impacts on downstream performance. First, we introduce our experimental conditions in Section \ref{sec:experiments setup}. Then, in Section~\ref{sec:are disentangled representations necessary}, we demonstrate that disentangled and general-purpose representations have similar performance, confirming the following experiment. Finally, we demonstrate how informativeness proves to be a stronger indicator of downstream performance in Section~\ref{sec:which property matters}.

\subsection{Experiments Setup} 
\label{sec:experiments setup}
We build upon the experiment conditions of \citet{van2019disentangled}.
Abstract visual reasoning tasks, i.e., RPMs, are solved through a two-stage process: data $\xrightarrow{\operatorname{Stage-1}}$ representations $\xrightarrow{\operatorname{Stage-2}}$ RPM answers.   We first train Stage-1 models in an unsupervised manner and evaluate their disentanglement and informativeness. Then, Stage-2 models are trained and evaluated on downstream tasks, yielding an abstract reasoning accuracy of a representation. Provided with a large amount of  $(\text{representation property score}, \text{downstream performance})$ pairs, we conduct a systematic study to investigate the necessity of disentanglement. More implementation details are available in Appendix A.

\textbf{Datasets.} We replicate the RPM generation protocol in \citet{van2019disentangled}. The panel images consist of disentanglement benchmark image datasets, namely, \textit{Abstract dSprites}~\citep{dsprites17,van2019disentangled} and \textit{3DShapes} \citep{3dshapes18}.  The rows of RPMs are arranged following the logical \textit{AND} of ground truth factors. As for hardness, we only reserve \textit{hard-mixed}, whose contexts and candidates are more confusing. According to the generation process, the size of generated RPMs is sufficiently large (about $10^{144}$), allowing us to produce fresh samples throughout training.

\textbf{Reference models.} Stage-1 models include 360 disentangled VAEs (denoted as DisVAEs) and 360 BYOLs, covering both disentangled and general-purpose representation learning methods. 
A diverse set of configurations are included. According to the histograms in Appendix C.4
, our choices of Stage-1 models span various disentanglement and informativeness scores. For Stage-2, we train 10 Stage-2 models (5 WReNs and 5 Transformers) for every Stage-1 model. These configurations are randomly sampled from a search space described in Appendix A.3 
and shared across Stage-1 models to ensure fair comparisons.

\textbf{Training protocol.} Training is conducted two-staged. Firstly, we train Stage-1 models in an unsupervised manner on the dataset consisting of RPMs' panels, i.e., \textit{Abstract dSprites} or \textit{3DShapes}. For DisVAE models, we use the training protocol of \citet{van2019disentangled}, while for BYOL models, we follow \citet{cao2022empirical}. In Stage-2, all models are trained for 10K iterations with a batch size of 32. After every 100 iterations, we evaluate the accuracy on newly generated 50 mini-batches of unseen RPM samples for validation and another 50 mini-batches for testing.

\textbf{Evaluation protocol.}  
We first evaluate the two stages separately. Then, we analyze the relationship between the two stages, i.e., representation properties and downstream performance. Specifically, to investigate the necessity of disentanglement, we are interested in whether more disentangled representations lead to better downstream performance. Further, can we find another metric that better accounts for downstream performance? Therefore,  for Stage-1,  we employ representation metrics described in Section~\ref{sec:representation_learning} to measure two aspects: disentanglement and informativeness. For all Stage-1 models, we compute the following metric scores: \textit{BetaVAE} score, \textit{FactorVAE} score, \textit{MIG}, \textit{SAP}, \textit{MED}, \textit{LR} accuracy, and negative normalized mean squared error of \textit{Linear Regression}. \textit{DCI Disentanglement} is only evaluated for DisVAEs.   Since we follow the tree-based implementation used in previous studies \citep{locatello2019challenging,van2019disentangled,dittadi2020transfer,trauble2021role} for easy comparison. We also provide \textit{DCI Disentanglement} score based on LASSO on both DisVAEs and BYOL (see Figure 9 and Appendix B.2).
For Stage-2, we inspect accuracy on newly generated test sets every 100 iterations, yielding accuracy for multiple training steps.  Since every step sees fresh samples, we employ training curves to measure sample efficiency following \citet{van2019disentangled, trauble2021role}. 

To summarize the downstream performance of a Stage-1 model, over 5 WReNs or 5 Transformers in Stage-2, we report the mean accuracy denoted as \awren{} or \atrans{}. Finally, we calculate the rank correlation (Spearman) between the mean performance of Stage-1 models (\awren{} and \atrans{}) at certain Stage-2 steps and their Stage-1 metric scores. Rank correlation has been widely adopted in the literature studying disentanglement downstream tasks \citep{van2019disentangled, locatello2019fairness, dittadi2020transfer, locatello2020weakly}. A larger correlation indicates a higher significance of the representation property on downstream performance.

\begin{figure*}[!t]
    \centering
    \begin{subfigure}{.48\textwidth}
        \centering
        \includegraphics[width=0.7\linewidth]{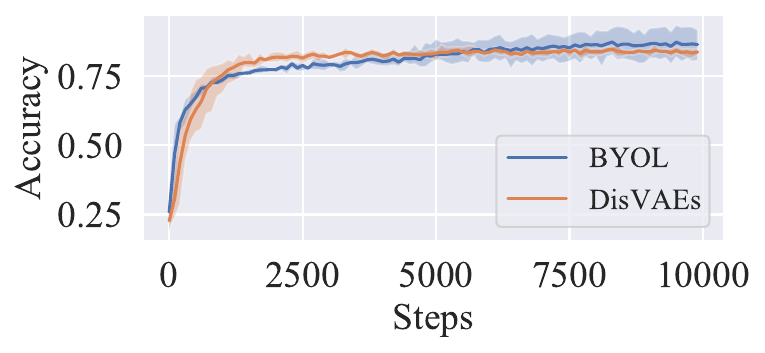}
        \caption{Stage-2=WReN}
        \label{fig:3dshapes_wren_acc_across_models}
    \end{subfigure}
    \begin{subfigure}{.48\textwidth}
        \centering
        \includegraphics[width=0.7\linewidth]{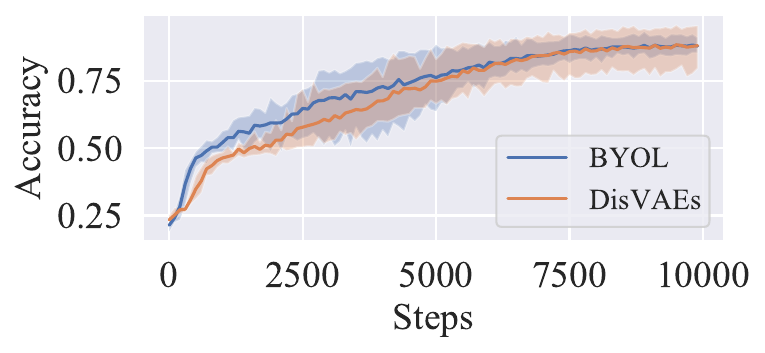}
        \caption{Stage-2=Transformer}
        \label{fig:3dshapes_trans_acc_across_models}
    \end{subfigure}
    \caption{Average test accuracy on \textit{3DShapes} throughout the training. The shaded area indicates the maximum and minimum values. We select the Stage-1 models with best \awren{} or \atrans{} among 3600 checkpoints on \textit{3DShapes}. Stage-1 models with disentanglement inductive bias (DisVAEs) are not necessarily better than those without such bias (BYOL) regarding sample efficiency and final accuracy. }
    \label{fig:3dshapes_acc_across_models}
\end{figure*}

\begin{table*}[!tb]
    \centering
    \begin{tabular}{c|l|c|c|c|c}
    \toprule
    Dataset & Stage1 & \awren{} & \atrans{} & MED (\awren{}) & MED (\atrans{}) \\
    \midrule
    \multirow{2}{*}{\textit{3DShapes}} & DisVAEs  & 84.8(0.91) & 87.0(6.36) & 0.60 & 0.68 \\
                                       & BYOL     & 87.1(4.68) & 88.0(2.62) & 0.13 & 0.12 \\
    \midrule
    \multirow{2}{*}{\makecell{\textit{Abstract}\\ \textit{dSprites}}} & DisVAEs    & 68.7{(1.39)} & 66.4{(7.06)} & 0.29 & 0.29 \\
                                       & BYOL     & 72.2{(3.11)} & 78.1{(1.75)} & 0.13 & 0.13 \\
    \bottomrule
    \end{tabular}
    \caption{Downstream test performance (\%) and MED scores of different Stage-1 models.  For each measurement (\awren{} and \atrans{}), we report the best among all Stage-1 models. The step with the highest validation accuracy is reported. The numbers in the parentheses are STDs of the 5 scores used in computing \awren{} or \atrans{}. The MED scores correspond to the checkpoints that achieve the reported \awren{} and \atrans{} performances.}
    \label{tab:final_acc}
\end{table*}

\subsection{Preliminary Experiments on Representation Variants  }
\label{sec:are disentangled representations necessary}
Previous studies have primarily focused on DisVAEs  \citep{van2019disentangled,locatello2019fairness,dittadi2020transfer,trauble2021role}.  In an endeavor to ensure the generalizability of our findings, two distinct representation variants are employed: disentangled representations and general-purpose representations.   In this section, we present the preliminary experiments conducted to compare the performance of these representations on the downstream task. Establishing comparable performance between these representation variants is important to substantiate the rationale behind considering both representation types in our analysis.

\textbf{Comparative Performance Analysis.} We show the downstream performance of different families of learning models described in Section~\ref{sec:experiments setup}, including disentanglement-oriented (DisVAEs) and general-purpose (DisVAEs) representations. To ensure equitable comparison, we select the most effective representations from a parameter search space of equal size (360 DisVAEs and 360 BYOLs). Each representation from Stage-1 was used to train 5 WReNs and 5 Transformers, and their performance was measured by averaging their downstream accuracy (\awren{}, \atrans{}). In this section, we present the models demonstrating the highest \awren{} and \atrans{} performance.

We first determine the final accuracy of disentangled representations versus less disentangled ones. In Table~\ref{tab:final_acc}, we present the \textit{best} results for \awren{} and \atrans{} achieved across various datasets. We select checkpoints to evaluate based on validation accuracy. It suggests that BYOL performs slightly better than DisVAEs.  In particular, the best \awren{} and \atrans{} of BYOL are higher than that of DisVAEs'. However, the MED scores of BYOL checkpoints in Table \ref{tab:final_acc} are significantly lower than those of DisVAEs, indicating that they are less disentangled.  Therefore, both entangled and disentangled representations can achieve similar performance, which suggests that it is valuable to investigate the necessity of disentanglement in general purpose and disentanglement methods.

\newcommand{\hone}{2.in}
\begin{figure*}[!t]
    \begin{subfigure}{.24\textwidth}
        \centering
        \includegraphics[height=\hone{}]{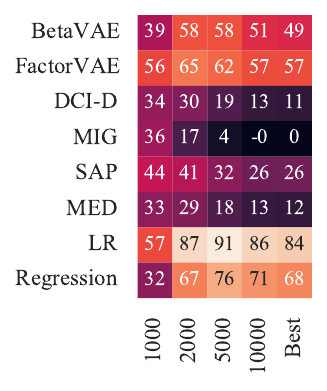}
        \caption{DisVAE-WReN}
        \label{fig:3dshapes_WReN_vae_heatmap}
    \end{subfigure}
    \begin{subfigure}{.24\textwidth}
        \centering
        \includegraphics[height=\hone{}]{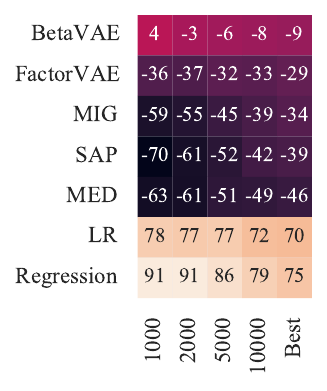}
        \caption{BYOL-WReN}
        \label{fig:3dshapes_WReN_byol_heatmap}
    \end{subfigure} 
    \begin{subfigure}{.24\textwidth}
        \centering
        \includegraphics[height=\hone{}]{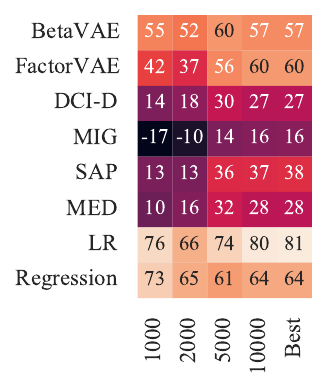}
        \caption{DisVAE-Transformer}
        \label{fig:3dshapes_transformer_vae_heatmap}
    \end{subfigure}
    \begin{subfigure}{.24\textwidth}
        \centering
        \includegraphics[height=\hone{}]{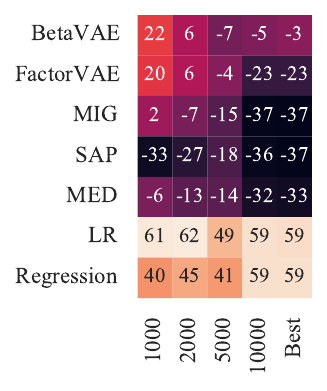}
        \caption{BYOL-Transformer}
        \label{fig:3dshapes_transformer_byol_heatmap}
    \end{subfigure}
    
    \caption{Rank correlations between \awren{} or \atrans{} and representation metrics on \textit{3DShapes}. We denote the step with the highest validation accuracy as ``Best". The brighter the panel, the more correlated the representation metric is with the downstream performance. }
    \vspace{-1.3em}
    \label{fig:3dshapes_heatmap}
\end{figure*}

\begin{figure*}[!t]
    \centering
    \includegraphics[width=\linewidth]{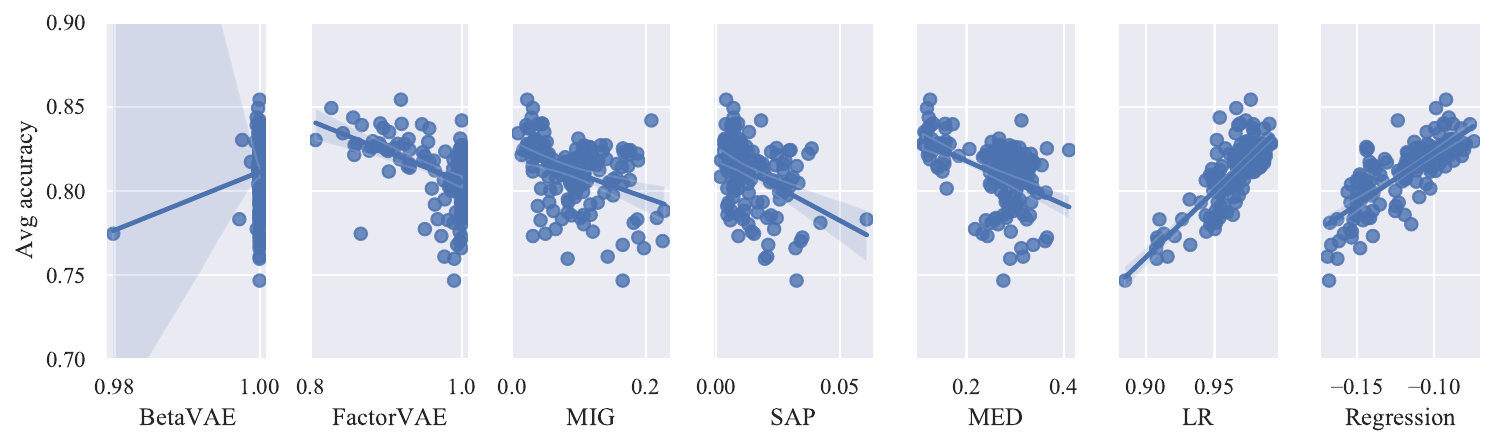}
    \vspace{-2em}
    \caption{Representation metrics versus \awren{} at step 10000, where Stage-1 models are BYOL, and the dataset is \textit{3DShapes}. We can observe a strong positive correlation between the informativeness metric scores and downstream accuracy. }
    \label{fig:3dshapes_wren_byol_step10000_metric_plot}
\end{figure*}

Now we examine another purported benefit of disentanglement: sample efficiency. According to the setting of Stage-2, every step sees fresh samples during training. Therefore, the number of training steps is a fraction of the total number of samples.  To gauge sample efficiency, we can observe how accuracy improves over training steps using training curves. This method was used by \citet{van2019disentangled} and \citet{trauble2021role}.
Figure~\ref{fig:3dshapes_acc_across_models} shows overviews of training trajectories of Stage-1 models with the highest performing \awren{} and \atrans{} on \textit{3DShpaes}.  For WReN as Stage-2 models (Figure~\ref{fig:3dshapes_wren_acc_across_models}), BYOL leads at the beginning, then DisVAEs catch up. Finally, BYOL converges at a higher accuracy. In contrast, when Stage-2 models are Transformers, BYOL's curve grows faster, but DisVAEs and BYOL converge with comparable performance. Generally, the two curves follow nearly identical patterns with small gaps, indicating no clear superiority in terms of sample efficiency between DisVAEs and BYOL. Consequently, disentanglement's enhancement of sample efficiency can be achieved with less disentangled representations. The same study is conducted on another set of general-purpose models, SimSiam \cite{simsiam}, where we reach similar results (see Appendix C.2).

\textbf{Ensuring Representation Diversity and Equivalency.}
Disentangled and general-purpose representation sets are of uniform size and have a broad spectrum of performance characteristics. Figure 12
 displays the metric scores, and Figure 13 
 and 14
 display the downstream performance of our selected DisVAEs and BYOL. We can see that both representations cover a wide range of metric scores and downstream accuracy (encompassing both superior and inferior ones).

\textbf{Summary:} The preliminary experiments reveal a comparable performance between disentangled and general-purpose representations on the abstract reasoning task. Moreover, both representation categories are of equal size and encapsulate a diverse performance range. This parity validates the subsequent evaluation of the correlation between disentanglement and downstream task performance across these representation variants.

\subsection{Informativeness as a Stronger Correlate Than Disentanglement}
\label{sec:which property matters}

To investigate the necessity of disentanglement, we analyze how various representation properties, including informativeness and disentanglement metrics, impact downstream performance. We first show that informativeness correlates most with downstream performance, indicating that disentanglement is not necessary. Further, we demonstrate that the previously claimed benefits of disentanglement \citep{bengio2013representation,higgins2016beta,van2019disentangled,locatello2019fairness,dittadi2020transfer}, are actually derived from its positive correlation with informativeness.


Recall that we train 720 Stage-1 and 7200 Stage-2 models (see Section~\ref{sec:experiments setup}). By taking \awren{} and \atrans{} as measurements (average reasoning accuracy over 5 WReNs or 5 Transformers), we yield 720 representations paired with their downstream performance. This section studies the importance of disentanglement and other representation properties. We must emphasize general trends and overall relationships to show different properties' significance. Therefore, we follow previous studies \cite{locatello2019challenging,van2019disentangled,locatello2019fairness,dittadi2020transfer,locatello2020weakly,trauble2021role} to analyze rank correlation (Spearman) between representation metric scores and downstream performance. If the correlation score is high, we can conclude that the representation property measured by the considered metric score is significant to downstream performance.

Our research builds upon previous studies \cite{locatello2019challenging,van2019disentangled,locatello2019fairness,dittadi2020transfer,locatello2020weakly,trauble2021role} by addressing two key issues. Firstly, we have corrected the previous underestimation of informativeness. Secondly, we have expanded the scope of both Stage-1 and Stage-2 models. Figure 8 
 illustrates how previous works yielded lower informativeness and resulted in lower correlations, leading to an overestimation of the importance of disentanglement. Additionally, the limited representation learning methods only confined to DisVAEs and reasoning models constrained to WReNs have restricted the generalizability of their results. By addressing these issues, our research leads to contrasting conclusions.

\newcommand{\htwo}{1.6in}
\begin{figure*}[!t]
    \begin{subfigure}{.48\textwidth}
        \centering
        \includegraphics[height=\htwo{}]{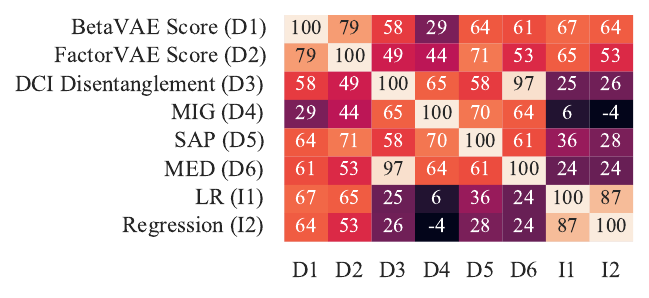}
        \caption{Overall rank correlations.}
        \label{fig:3dshapes_WReN_vae_self_heatmap}
    \end{subfigure}
    \begin{subfigure}{.48\textwidth}
        \centering
        \includegraphics[height=\htwo{}]{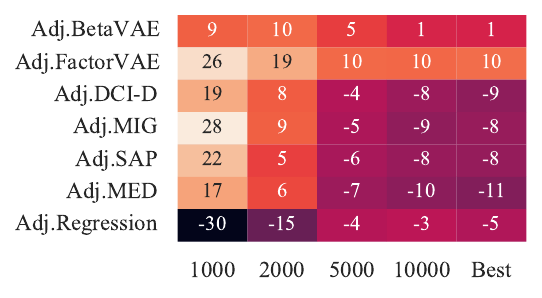}
        \caption{Correlations of adjusted metrics.}
        \label{fig:3dshapes_WReN_vae_adj_heatmap}
    \end{subfigure}
    \vspace{-0.8em}
    \caption{(a) Correlations between metrics and (b) correlations between adjusted metrics and downstream accuracy when using DisVAEs-WReN pipeline on \textit{3DShapes}. Disentanglement metrics exhibit positive correlations with informativeness. Yet when conditioned on close \keyfact{}, their adjusted versions show mild correlations.  }
    \vspace{-1.5em}
    \label{fig:3dshapes_self_adj_heatmap}
\end{figure*}

\textbf{The more significant representation property. } 
We calculate the rank correlation between downstream accuracy with disentanglement and informativeness scores.  Meanwhile, we report rank correlation at steps 1K, 2K, 5K, and 10K, and the step with the highest validation accuracy. Because each step involves new samples, the training steps represent a fraction of the sample size. By examining correlations at various training steps, we can determine the impact of representation properties on sample efficiency.  

Figure~\ref{fig:3dshapes_heatmap} displays rank correlations between representation metric scores and abstract reasoning test accuracy on \textit{3DShapes}. Firstly, at each step throughout the training process, there is always at least one informative metric, such as \textit{Logistic Regression} accuracy (LR) or negative error of linear regressions, that exhibits the most significant correlation with downstream performance. Specifically, LR always correlates more positively than disentanglement metrics.
The strong correlation is exploited for all considered models at multiple steps. Since LR and linear regression require sufficient information to be captured and extracted easily from representations, we can conclude that the \keyfact{} matters most in broad conditions. In contrast, we observe that the importance of disentanglement varies among Stage-1 model families. Disentangled representation learning models (DisVAEs) exhibit strong positive correlations for several disentanglement metrics (but weaker than at least one of \keyfact{} metrics), such as \textit{FactorVAE} score and \textit{DCI Disentanglement}. However, their significance does not apply to BYOL, where the correlation of disentanglement is mild or even negative. In Figure~\ref{fig:3dshapes_wren_byol_step10000_metric_plot}, we plot the (\awren{}, metric score) pairs at step 10000.  Indeed, for BYOL-WReN on \textit{3DShapes}, we can see the reg-plot provides a good fit of downstream accuracy and informativeness metrics. As for disentanglement metrics, we can see that \textit{BetaVAE} score and \textit{FactorVAE} score suffer from narrow spreads. The regression lines have negative slopes for \textit{MIG}, \textit{SAP}, and \textit{MED}.  We conduct a similar analysis on another dataset (\textit{Abstract dSprites}), and another Stage-1 model (SimSiam), and take the same observations. Please refer to Appendix C.4
and Appendix C.2.
On the fairness downstream task \cite{locatello2019fairness}, we also find that informativeness correlates most (see Appendix C.1 and Table 3).

\textbf{Summary:} The importance of \keyfact{} surpasses that of disentanglement, as evidenced by consistent results across various datasets and model structures.

\textbf{Understanding for the previously claimed success of disentanglement.} Previous works \citep{van2019disentangled,locatello2019fairness,dittadi2020transfer,locatello2020weakly} have reported empirical evidence backing up the advantages of disentangled representations. Consistently, we observe relatively strong correlations with disentanglement metrics, especially when Stage-1 models are DisVAEs in Figure~\ref{fig:3dshapes_heatmap}. Based on our conclusion on the significance of the \keyfact{}, we study the DisVAE-WReN case. We provide some insights to explain why the disentanglement metrics have a high correlation to downstream performance in some cases.

The correlations between various metrics have been computed and the results are displayed in Figure~\ref{fig:3dshapes_WReN_vae_self_heatmap}. Upon analysis of DisVAEs, it was discovered that there is a strong correlation between informativeness and disentanglement. Additionally, \keyfact{} has a significant correlation with both \textit{FactorVAE} score and \textit{BetaVAE} score. This is evident in Figure~\ref{fig:3dshapes_WReN_vae_heatmap} where these disentanglement metrics have a strong correlation with downstream performance. However, other disentanglement metrics have only a mild correlation with \keyfact{} and are ineffective for downstream performance. Therefore, it can be concluded that disentanglement metrics cannot truly predict downstream performance, but \keyfact{} can.

To ``purify'' the effect of disentanglement, a natural question is: If two representations are of close informativeness, is the more disentangled one more helpful for downstream tasks? For this, we employ adjusted metrics in \citet{locatello2019fairness}: $\texttt{Adj. Metric} = \texttt{Metric} - \frac{1}{5}\sum_{i \in N(\text{LR})} \texttt{Metric}_{i}$.
For a representation and a certain metric (we care more about disentanglement metrics), we denote its original metric score as $\texttt{Metric}$. Then we find its 5 nearest neighbors in terms of LR, which we write as $N(\text{LR})$. Finally, the difference between the original metric score and the mean score of the nearest neighbors is reported as adjusted metrics. Intuitively, we calculate the relative disentanglement for representations with close LR.

Figure~\ref{fig:3dshapes_WReN_vae_adj_heatmap} displays correlations between adjusted metrics and downstream performance. We can find that all adjusted disentanglement metrics correlate mildly with downstream performance. From this, we can see that when informativeness is close, being disentangled contributes only a small portion to the downstream performance when the downstream training steps are limited (In our case, less than or equal to 2000 steps, see Figure~\ref{fig:3dshapes_acc_across_models} and Figure~\ref{fig:3dshapes_self_adj_heatmap}).  

\textbf{Summary:} The \keyfact{} metric is more reliable in predicting downstream performance. Disentanglement provides only marginal additional advantages at the outset of downstream training.
\section{Conclusion}

In this paper, we demonstrate that dimension-wise disentanglement is not necessary for abstract visual reasoning.  We identify that informativeness is of the most significance for downstream performance. Informativeness explains the previously claimed benefits of disentanglement.  As abstract reasoning is a fundamental and indicative task, our study could have significant implications for a range of tasks.

\section*{Acknowledgments}
This work is supported by the Ministry of Science and Technology of the People's Republic of China, the 2030 Innovation Megaprojects ``Program on New Generation Artificial Intelligence'' (Grant No. 2021AAA0150000).  This work is also supported by the National Key R\&D Program of China (2022ZD0161700).

\bibliography{aaai24}
\newpage
\appendix
\section{Reproducibility}
\label{sec:reproducibility}
In this Section, we provide implementation details to ensure reproducibility.  All experiments are run on a machine with 2 Intel Xeon Gold 5218R 20-core processors and 4 Nvidia GeForce RTX 3090 GPUs.

\subsection{Representation Learning Methods}
We include both disentangled representation learning methods and general-purpose representation learning methods. i.e., DisVAEs and BYOL \citep{grill2020bootstrap}.

\textbf{DisVAEs implementation.}  The DisVAEs include $\beta$-VAE~\citep{higgins2016beta}, AnnealedVAE~\citep{burgess2018understanding}, $\beta$-TCVAE~\citep{chen2018isolating}, FactorVAE~\citep{kim2018disentangling}, and DIP-VAE-I and DIP-VAE-II~\citep{kumar2017variational}. We use the output of the encoder, the mean of $q_{\phi}(z|x)$, as representations. Hereafter, we introduce details for each method. The above methods encourage disentanglement by adding regularizers to ELBO. Adopting the notation in \citet{tschannen2018recent}, their objectives can be written in the following unified form: 
\begin{align*}
&\mathbb{E}_{p(x)}[\mathbb{E}_{q_{\phi}(z|x)}[- \log p_{\theta}(x|z)]] \\
&\quad + \lambda_1\mathbb{E}_{p(x)}[R_1(q{\phi}(z|x))] + \lambda_{2}R_2(q_{\phi}(z)),
\end{align*}
where $q_{\phi}(z|x)$ is the posterior parameterized by the output of the encoder, $p_{\theta}(x|z)$ is induced by the decoder output, $R_1, R_2$ are the regularizer applying to the posterior and aggregate posterior, and $\lambda_1, \lambda_2$  are the coefficients controlling regularization. In the objective of $\beta$-VAE, $\beta = \lambda_1 > 1, \lambda_2=0$. Taking $R_1(q_{\phi}(z|x)) := D_{KL}[q_{\phi}(z|x) || p(z) ]$  forces the posterior to be close to the prior (usually unit gaussian), hence penalizing the capacity of the information bottleneck and encourage disentanglement. FactorVAE and $\beta$-TCVAE takes $\lambda_1=0, \lambda_2=1$. With $R_2(q_{\phi}(z)) := TC(q_{\phi}(z))$, they penalize the Total Correlation (TC) \citep{watanabe1960information}. FactorVAE estimates TC by adversarial training, while $\beta$-TCVAE estimates TC by biased Monte Carlo sampling. Finally, DIP-VAE-I and DIP-VAE-II take $\lambda_1=0,\lambda_2 \geq 1$ and $R_2(q_{\phi}(z)) := || \textnormal{Cov}_{q_{\phi}(z)} - I||^{2}_{F}$, penalizing the distance between aggregated posterior and factorized prior. 

We use the code and configurations from the DisLib \footnote{\url{https://github.com/google-research/disentanglement_lib.git}} \citep{locatello2019challenging}. 
As for parameters, we use the same sweep as \citet{van2019disentangled}: for each one of the 6 DisVAEs, we use 6 configurations. We train each model using 5 different random seeds. Since we consider 2 datasets (\textit{3DShapes} and \textit{Abstract dSprites}), 
 finally, we yield $6*6*5*2=360$ DisVAE checkpoints.

\begin{figure*}[!th]
    \centering
    \includegraphics[width=\textwidth]{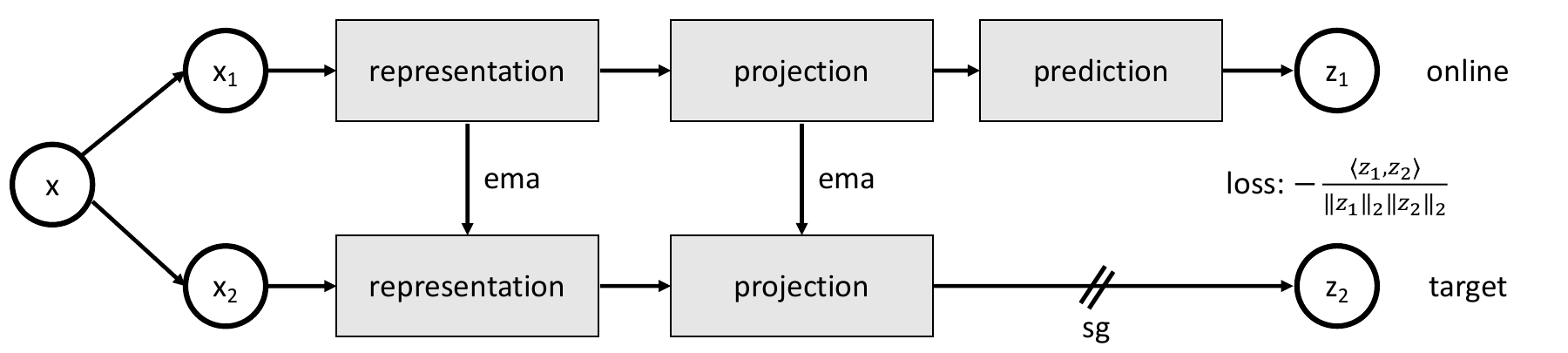}
    \caption{The pipeline of BYOL \citep{grill2020bootstrap}.}
    \label{fig:byol}
\end{figure*}
\textbf{BYOL implementation.} BYOL \citep{grill2020bootstrap} is a contrastive learning method. Figure~\ref{fig:byol} shows its pipeline. For each image $x$, we first create two ``views" of it by data augmentation, i.e., $x_1$ and $x_2$. Then they are input to the siamese encoders: the online encoder and the target encoder. Specifically, $x_1$ is fed to the online encoder, while $x_2$ is fed to the target encoder, yielding the output $z_1$ and $z_2$, respectively. As for architectures, both encoders share the same representation network and projection MLP. The prediction MLP is appended to the online encoder in order to avoid BYOL learning trivial representations. The objective of BYOL is
\newcommand{\norm}[1]{\ensuremath{\|#1\|}}
\begin{equation}
    \mathcal{L}=-\frac{\langle z_1, z_2 \rangle}{\norm{z_1}_2 \norm{z_2}_2}.
\end{equation}
We are pulling the representations of the two ``views" close. While training, the online encoder's parameters are updated by gradient descent. However, the target encoder's parameters are updated by the online parameters' Exponential Moving Average (EMA) \citep{he2020momentum}. After training, we only keep the online encoder and use the output of the representation network as representations. 
\begin{table}[!tb]
\centering
\begin{tabular}{l} 
     \toprule
     \textbf{Representation Network} \\
     \midrule
     \textbf{input}:  $64\times64$ images \\ 
     \textbf{pipeline}:  \\
     \quad \quad \quad 4$\times$4 conv, stride 2, 32-channel \\
     
     \quad \quad \quad 4$\times$4 conv, stride 2, 32-channel\\
    
     \quad \quad \quad 4$\times$4 conv, stride 2, 64-channel\\

     \quad \quad \quad 4$\times$4 conv, stride 2, 64-channel\\
     
     \quad \quad \quad 4$\times$4 conv, stride 2, 128-channel\\
    
     \quad \quad \quad 1$\times$1 conv, stride 1, $D$-channel\\
     
     \bottomrule
\end{tabular}
\caption{The representation network architecture of our BYOL implementation, following \citet{cao2022empirical}.  Besides, there is a ReLU activation layer and a possible normalization layer following each convolutional layer to create a stack of (Conv-ReLU-Norm) blocks. The normalization stratege $norm$ and representation dimension $D$ are parameters to be set.}
\label{tab:encoder}
\end{table}

We use the PyTorch implementation of BYOL \footnote{\url{https://github.com/lucidrains/byol-pytorch.git}}. We use the representation network architecture as shown in Table~\ref{tab:encoder}, where the representation dimension $D$ is a parameter to be set. Except for normalization and output dimensions, the representation network architecture of BYOL and the encoder architecture of DisVAEs are similar. As for predictor and projector, we use the pipeline Linear$\rightarrow$ BN $\rightarrow$ ReLU $\rightarrow$ Linear with 256 hidden neurons. We train the BYOLs for 105 epochs using the Adam optimizer with $\beta_1=0.9$, $\beta_2=0.999$, $\epsilon=10^{-8}$, and learning rate ($lr$) as a variable parameter. For augmentation, we use the pipeline of \citet{cao2022empirical} (in PyTorch-style): 
\begin{enumerate}
    \item \textit{RandomApply(transforms.ColorJitter($x_{\text{jit}}$, $x_{\text{jit}}$, $x_{\text{jit}}$, 0.2),~p=0.8)}
    \item \textit{RandomGrayScale(p=$p_{\text{gray}}$)}
    \item \textit{RandomHorizontalFlip()}
    \item \textit{RandomApply(transforms.GaussianBlur((3,3),~(1.0, 2.0)), p=0.2)}
    \item \textit{RandomResizeCrop(size=(64, 64), scale=($x_{\text{crop}}$,~1.0))}
\end{enumerate}
The $x_{\text{jit}}$, ~$p_{\text{gray}}$, ~and $x_{\text{crop}}$ are parameters to be set. $x_{\text{jit}}$ controls how much to jitter brightness, contrast, and saturation. $p_{\text{gray}}$ controls the probability to convert the image to grayscale. $x_{\text{crop}}$ defines  the lower bound for the random area of the crop.

We perform a parameter sweep on the cross product of intervals of parameters $D$, $norm$, $lr$, $x_{\text{jit}}$,~$p_{\text{gray}}$, and $x_{\text{crop}}$. On \textit{3DShapes}, we use the following parameter grid (in scikit-learn style): 
\lstset{basicstyle=\ttfamily\footnotesize,breaklines=true}
\begin{lstlisting}[language=Python]
[
  {'D': [32, 64, 128], 'lr': [3e-2, 3e-3], 'norm': [BatchNorm()], 
  'x_jit': [0.6, 0.8], 'p_gray': [0.5, 0.7, 0.9], 'x_crop': [1.0]},
  {'D': [256], 'lr': [3e-4, 3e-5], 
  'norm': [BatchNorm(), GroupNorm(num_groups=4)], 'x_jit': [0.4, 0.8], 
  'p_gray': [0.3, 0.5, 0.7], 'x_crop': [1.0]}
]
\end{lstlisting}
On \textit{Abstract dSprites}, we use the following parameter grid:
\begin{lstlisting}[language=Python]
[
  {'D': [32, 64, 128], 'lr': [3e-3, 3e-4], 'norm': [BatchNorm()], 
  'x_jit': [0.6, 0.8], 'p_gray': [0.0, 0.1, 0.2], 'x_crop': [0.6]},
  {'D': [256], 'lr': [3e-4, 3e-5], 
  'norm': [BatchNorm(), GroupNorm(num_groups=4)], 'x_jit': [0.4, 0.8],
  'p_gray': [0.0, 0.1, 0.2], 'x_crop': [0.6]}
]
\end{lstlisting}
For each parameter configuration, we run it with 3 random seeds. Finally, we trained 360 BYOLs in total.

\subsection{Abstract Reasoning Methods}
\label{sec:appendix_resoning_methods}
We include two abstract reasoning network architectures: WReN \citep{barrett2018measuring,van2019disentangled} and Transformer \citep{vaswani2017attention,hahne2019attention}. 

\textbf{WReN implementation.} WReN consists of two parts: graph MLP and edge MLP. Here we use the same notations as in Section~\ref{sec:background_reasoning_methods}. For the representations of a trial $\operatorname{Stage1}(T_i)$, edge MLP takes a pair of representations in $\operatorname{Stage1}(T_i)$ as input and embed them to edge embeddings. Then all edge embeddings of $\operatorname{Stage1}(T_i)$ (in total $C_9^2$=36) are added up and input to the graph MLP. Finally, the graph MLP output a scalar score, predicting the correctness of the trial $T_i$.  

We use the code \citep{van2019disentangled} to implement WReN. And we use the same parameter searching spaces as them. All WReNs are trained in 10K steps with a batch size of 32. The learning rate for the Adam optimizer is sampled from the set $\{0.01, 0.001, 0.0001\}$ while $\beta_1=0.9$, $\beta_2=0.999$, and $\epsilon=10^{-8}$.
For the edge MLP  in the WReN model, we uniformly sample its hidden units in 256 or 512, and we uniformly choose its number of hidden layers in 2, 3, or 4.
Similarly, for the graph MLP  in the WReN model, we uniformly sample its hidden units in 128 or 512, and we uniformly choose its number of hidden layers in 1 or 2 before the final linear layer to predict the final score.
We also uniformly sample whether we apply no dropout, dropout of 0.25, dropout of 0.5, or dropout of 0.75 to units before this last layer.

\textbf{Transformer implementation.} We simplify the architecture of \citet{hahne2019attention}. Here we treat $\operatorname{Stage1}(T_i)$ as a sequence. We first linear project all representations  and prepend them with a learnable \texttt{[class]} token. We add them with learnable positional embeddings. Then they are input into a stack of Transformer blocks \citep{vaswani2017attention}. Finally, an MLP predicts a scalar score from the class embedding of the final Transformer block.

We implement the Transformer architecture ourselves with utilities of the DisLib code base. All Transformers are trained for the same steps and same batch size as WReN, i.e., 10K steps with a batch size of 32. We use the Adam optimizer with weight decay and cosine learning rate scheduler. The learning rate for the Adam optimizer is uniformly selected from $\{5e-4, 6e-4, 7e-4\}$. The depth of Transformer blocks is uniformly set to be 2, 3, or 4. The dimensions of $q,k,v$ of the self-attention model are uniformly 32 or 64. The MLP head uses the same architecture and parameter space as the graph MLP in WReN. For other fixed parameters, please refer to our codes for details.

\subsection{Representation Metrics}
\label{sec:appendix_representation_metrics}
In the main text, we employ disentanglement and informativeness metrics to measure the properties of representations. Here we provide more details. 

\textbf{Disentanglement metrics.} We use the setup and implementation of \citet{locatello2019challenging}. Here we briefly introduce the details of our considered metrics. Namely,  \textit{BetaVAE} score~\citep{higgins2016beta},  \textit{FactorVAE} score  ~\citep{kim2018disentangling},  \textit{Mutual Information Gap} ~\citep{chen2018isolating} , \textit{SAP} ~\citep{kumar2017variational},  and \textit{DCI Disentanglement} ~\citep{eastwood2018framework}. We also include \textit{MED} \citep{cao2022empirical}. The \textit{BetaVAE} score and the \textit{FactorVAE} score predict the intervened factor from representations to measure disentanglement. The \textit{Mutual Information Gap} and \textit{SAP} compute the gap in response for each factor between the two highest representation dimensions. The difference is that MIG measures mutual information while SAP measures classification accuracy. The \textit{DCI Disentanglement} calculates the entropy of the relative importance of a latent dimension in predicting factors. {We follow previous studies \citep{locatello2019challenging, van2019disentangled, locatello2019fairness, dittadi2020transfer} to develop a Gradient Boosting Tree (GBT) for prediction during the \textit{DCI Disentanglement} evaluation. Though according to \citet{eastwood2018framework} any classifier could be used.  } As reported by \citet{cao2022empirical}, the GBT takes hours to train from high-dimensional representations learned by BYOL. Thus we only report \textit{DCI Disentanglement} score for DisVAEs. 
\textit{MED} is a modified version of \textit{DCI Disentanglement}. Unlike \textit{DCI Disentanglement}, which measures importance based on classifier weights, MED measures importance through Mutual Information. This approach eliminates the need to develop classifiers and allows for fair comparison across different representation dimensions \citep{cao2022empirical}.

\textbf{Informativeness metrics.} We evaluate the informativeness of our representations using LR and linear regression following \citet{van2019disentangled} and \citet{eastwood2018framework}. Our method entails training a \textit{Logistic Regression} model to predict, or a linear regression model to regress, factor values from the representations, utilizing 10000 training samples \citep{locatello2019challenging,van2019disentangled, eastwood2018framework}. We then measure the accuracy of LR and the negative normalized mean squared error (NMSE) of regression. These scores are averaged across all generative factors for a given representation.


\subsection{Abstract Visual Reasoning Datasets}
\label{sec:appendix_abstract_visual_reasoning_datasets}
We use the two abstract visual reasoning datasets developed by \citet{van2019disentangled}. i.e., Ravens' Progressive Matrices created from \textit{3DShapes} \citep{3dshapes18} and \textit{Abstract dSprites} \citep{dsprites17, van2019disentangled}. 

\begin{figure}[ht]
    \begin{subfigure}{.48\textwidth}
        \centering
        \includegraphics[width=0.8\textwidth]{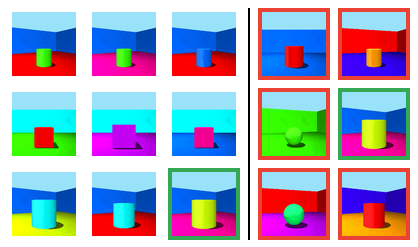}
        \caption{}
        \label{fig:3dshapes_solution}
    \end{subfigure}
    \begin{subfigure}{.48\textwidth}
        \centering
        \includegraphics[width=.8\textwidth]{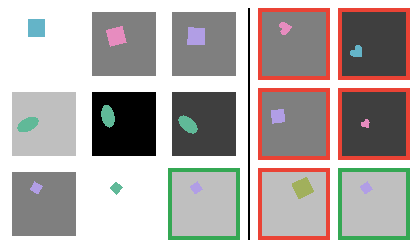}
        \caption{}
        \label{fig:dsprites_solution}
    \end{subfigure}
    \caption{RPM questions with solutions on (a) \textit{3DShapes} and (b) \textit{Abstract dSprites}. }
    \label{fig:reasoning_question_with_answers}
\end{figure}

We sketch the rules here by taking the RPM in Figure~\ref{fig:3dshapes_question} as an example.  The reasoning attributes are the ground truth factors of \textit{3DShapes}. i.e., floor hue, wall hue, object hue, scale, shape, and orientation.
In the $3\times 3$ matrix, each row has 1, 2, or 3 fixed ground truth factors. The 3 rows share the same fixed ground truth factors but with different values. By examining the context panels, one can discover the logical relationship between the factors. The task is to fill in the missing panel with one of the candidate options. In the case of Figure~\ref{fig:3dshapes_question}, we can infer from the context that the fixed factors are wall hue, shape, and orientation. Using the first 2 panels in the third row, we know that the wall hue is blue, the shape is cylinder, and the orientation is the azimuth that makes the wall corner appear on the right side of the image. Based on these factor values, we choose the candidate with the closest match, which is shown in Figure~\ref{fig:3dshapes_solution}. Additionally, Figure~\ref{fig:dsprites_solution} provides a sample of RPMs with answers for \textit{Abstract dSprites}.

\section{Details and Comparisons of Empirical Design}
This section provides a comprehensive analysis of the empirical designs employed in our study. We commence by conducting a meticulous comparison of various measurements of informativeness, elucidating the rationale behind our selection of specific metrics. Subsequently, we present the outcomes of diverse implementations of the DCI-Disentanglement metric \cite{eastwood2018framework}, robustly establishing the consistency of our conclusions across different approaches. Furthermore, we present results about different grouping methodologies employed for the Stage-1 models. Whether considering all Stage-1 methods as a unified group or categorizing them based on their inductive bias, our findings consistently underscore the paramount importance of informativeness.

\newcommand{\hlr}{10.5em}
\begin{figure*}[!th]
    \begin{subfigure}{.7\textwidth}
        \centering
        \includegraphics[height=\hlr{}]{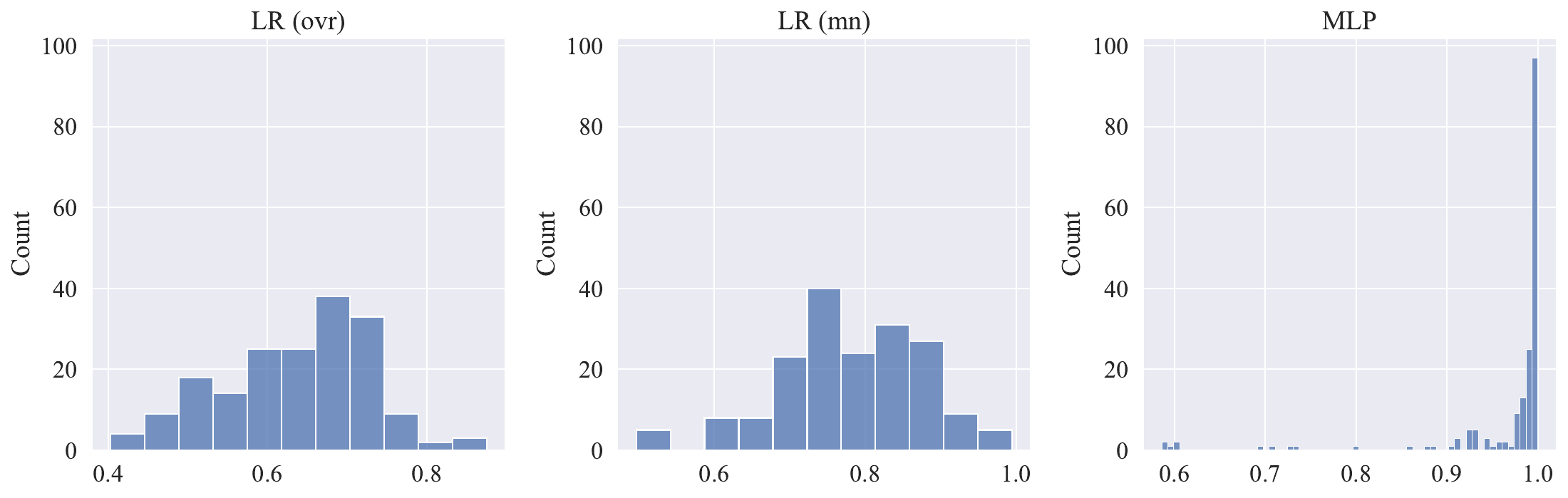}
        \caption{}
        \label{fig:3dshapes_LR_hist}
    \end{subfigure}
    \begin{subfigure}{.28\textwidth}
        \centering
        \includegraphics[height=\hlr{}]{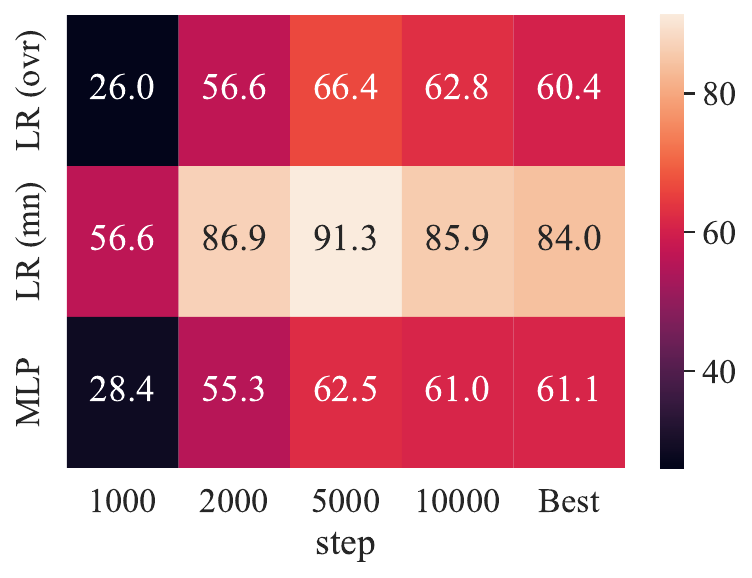}
        \caption{}
        \label{fig:3dshapes_LR_heat_map}
    \end{subfigure}
    \vspace{-1em}
    \caption{(a) Prediction accuracy by ``multinomial'' LR, denoted as LR (mn), ``one v.s. rest'' LR, denoted as LR (ovr), and MLP of DisVAEs on \textit{3DShapes}. (b) LR (mn), LR (ovr), and MLP's correlations with the downstream performance of the DisVAEs-WReN pipeline.}
    \label{fig:3dshapes_LR}
\end{figure*}

\newcommand{\hdci}{11.5em}
\begin{figure*}[th]
    \centering
    \begin{subfigure}{.48\textwidth}
        \centering
        \includegraphics[height=\hdci{}]{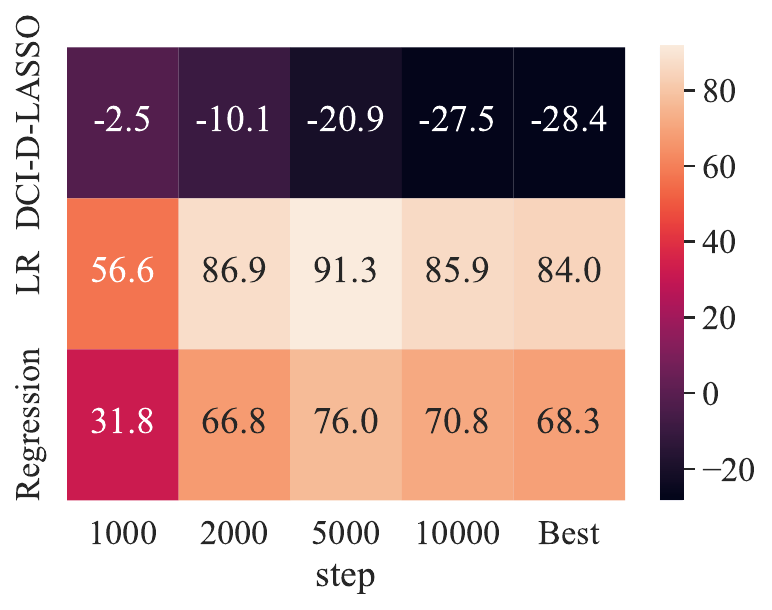}
        \caption{DisVAEs-WReN}
        \label{fig:dci_heat_vae}
    \end{subfigure}
    \begin{subfigure}{.48\textwidth}
        \centering
        \includegraphics[height=\hdci{}]{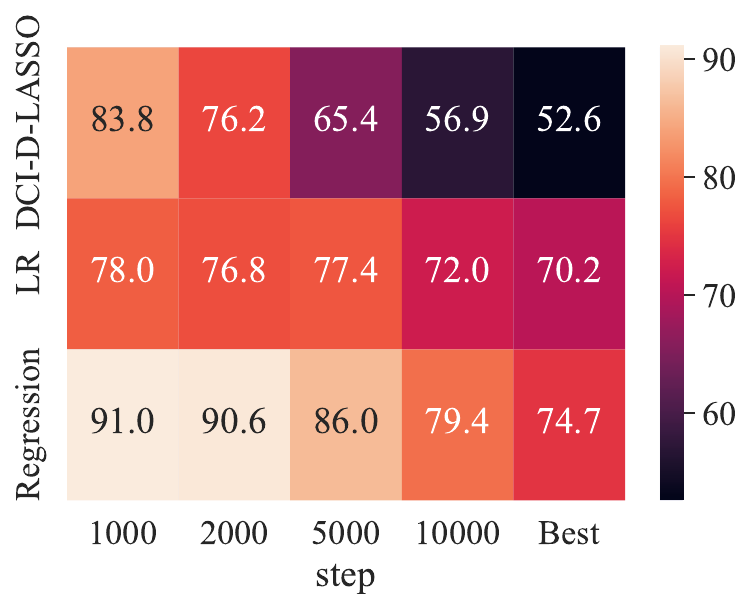}
        \caption{BYOL-WReN}
        \label{fig:dci_heat_byol}
    \end{subfigure}
    \caption{Correlations with DCI-D-LASSO on \textit{3DShapes}.}
    \label{fig:dci_ablation}
\end{figure*}

\newcommand{\hall}{11.5em}
\begin{figure*}[htb]
    \centering
    \begin{subfigure}{.48\textwidth}
        \centering
        \includegraphics[height=\hall{}]{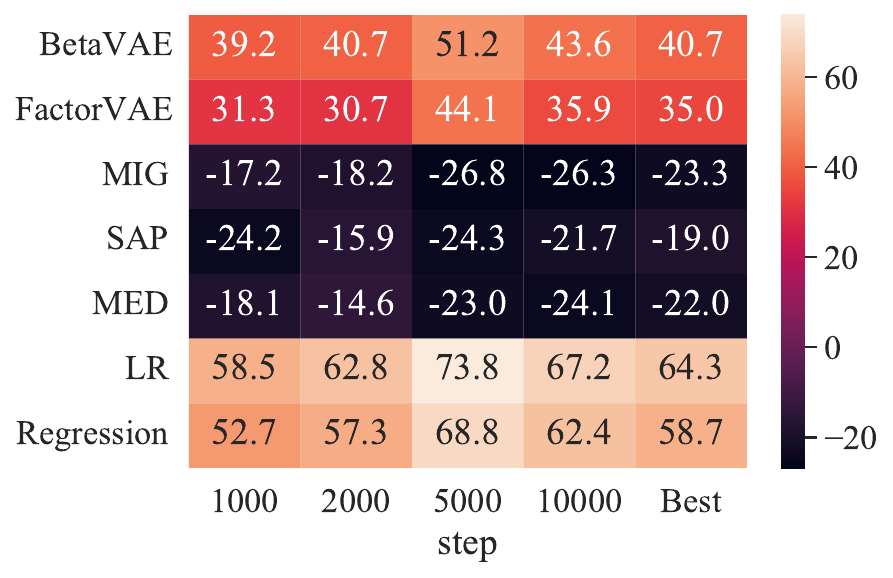}
        \caption{All Stage-1-WReN}
    \end{subfigure}
    \begin{subfigure}{.48\textwidth}
        \centering
        \includegraphics[height=\hall{}]{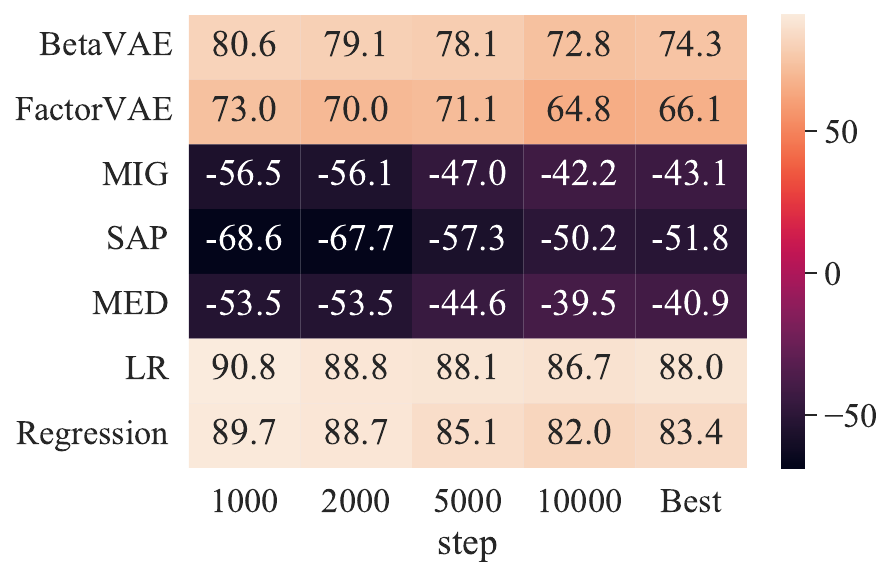}
        \caption{All Stage-1-Transformer}
    \end{subfigure}
    \caption{Correlations when considering all Stage-1 models as one set on \textit{3DShapes}.}
    \label{fig:heatmap_one_set}
\end{figure*}

\subsection{Measurement of Informativeness}
\label{sec:informativeness_discussion}
Informativeness not only necessitates the adequate encoding of factors in representations but also underscores the ease-of-use \cite{eastwood2018framework,eastwood2022dci}. Therefore, it is crucial to consider the capability of the prediction or regression model. In this context, we specifically classify the generative factors as an informativeness task. We compare three classification models: ``one v.s. rest'' LR, ``multinomial'' LR, and MLP. 

Unlike \citet{van2019disentangled}, we use a ``multinomial'' classification scheme instead of the ``one v.s. rest'' approach for multi-class classification. As depicted in Figure~\ref{fig:3dshapes_LR_hist}, prediction accuracy is higher with the ``multinomial'' LR model than with the ``one v.s. rest'' LR model for the same set of representations. Additionally, Figure~\ref{fig:3dshapes_LR_heat_map} shows that the correlation between ``one v.s. rest'' LR scores and downstream performance are weaker. 
On the other hand, if the classification model is too capable, like MLP, accuracy will fail to differentiate representations (as shown in Figure \ref{fig:3dshapes_LR_hist}). Consequently, we can see in Figure \ref{fig:3dshapes_LR_heat_map} that the correlations are falsely low.
Therefore, underestimating or overestimating the informativeness score can lead to underestimating its impact on downstream performance. To ensure a more accurate estimation of informativeness, we use the ``multinomial'' LR model as a measurement.

\subsection{Implementation of DCI-Disentanglement}
\label{sec:dci_discussion}
We utilized the DCI-Disentanglement (DCI-D) implementation, employing Gradient Boost Trees (GBT), in accordance with previous studies \citep{van2019disentangled, locatello2019challenging, locatello2019fairness}. Nonetheless, the higher representation dimensions of BYOL require hours to develop GBTs. Therefore, we have not reported GBT-based DCI-D scores for BYOL. For completeness, here we adopt an alternative implementation, DCI-D-LASSO \cite{eastwood2018framework}, which computes the DCI-D score using the weight of LASSO as an importance matrix. This approach is feasible for BYOL.

In Figure \ref{fig:dci_ablation}, we can see that DCI-D-LASSO correlates less than informativeness scores for both Stage-1 methods, suggesting that our conclusion still holds despite the implementation of DCI-D.

\subsection{Grouping of Stage-1 to Calculate Correlation}
The main text provides a detailed view of the significance of representation metrics by showing correlation results on DisVAEs and BYOL separately. For a more holistic comprehension, we have consolidated all the trained Stage-1 models into one singular set. The heat-map depicted in Figure \ref{fig:heatmap_one_set} manifests that informativeness persists as the most correlated aspect, thereby ensuring the credibility of our results.

\newcommand\hzero{10.0em}

\section{Additional results}
\label{sec:additional_results}
\subsection{Correlation Results on Fairness Task}
\label{sec:fairness}
\begin{table*}[bth]
    \centering
    \begin{tabular}{l|c|c|c|c|c|c|c|c}
        \toprule
        Stage-1 & BetaVAE & FactorVAE & DCI-D & MIG & SAP & MED & LR & Regression \\ \midrule
        DisVAEs & -64.02 & -60.35 & -26.37 & -2.57 & -31.71 & -26.11 & -95.90 & -85.82 \\ 
        BYOL & -12.30 & 27.81 & — & 39.76 & 52.13 & 48.95 & -97.03 & -84.17 \\ \bottomrule
    \end{tabular}
    \caption{Correlation between unfairness score and representation metrics on \textit{3DShapes}.}
    \label{tab:fairness}
\end{table*}
Following \cite{locatello2019fairness}, we assume that there exists a target variable, denoted as $\rvy$, that requires prediction using a representation. However, the sensitive variable, $\rvs$, remains unobservable.  For each trained Stage-1 model, we consider every possible combination of factors of variation as both target and sensitive variables. For the prediction of the target variable, we consider using Logistic Regression (LR), which was trained on 10000 labeled examples. To measure the unfairness of prediction, we use the following score.
\begin{align*}
    \texttt { unfairness }(\hat{\mathbf{y}})=\frac{1}{|S|} \sum_s T V(p(\hat{\mathbf{y}}), p(\hat{\mathbf{y}} \mid \mathbf{s}=s)) \forall y.
\end{align*}
where $T V$ is the total variation. The unfairness score for each trained representation is the average unfairness of all downstream classification tasks we considered for that representation.

We calculate the unfairness scores of our trained encoders on the \textit{3DShapes} dataset. As shown in Table \ref{tab:fairness}, we computed the correlation between unfairness score and downstream performance, as we did for abstract reasoning. We can see that informativeness metrics correlate the most negatively (note that the lower the unfairness score, the better). We reached the same conclusion as in the abstract reasoning task: disentanglement is not necessary, and informativeness correlates most among all considered metrics.
\subsection{Additional Results on SimSiam}
\label{sec:appendix_ablation_simsiam}
\begin{figure}[!t]
    \centering
    \includegraphics[width=0.4\textwidth]{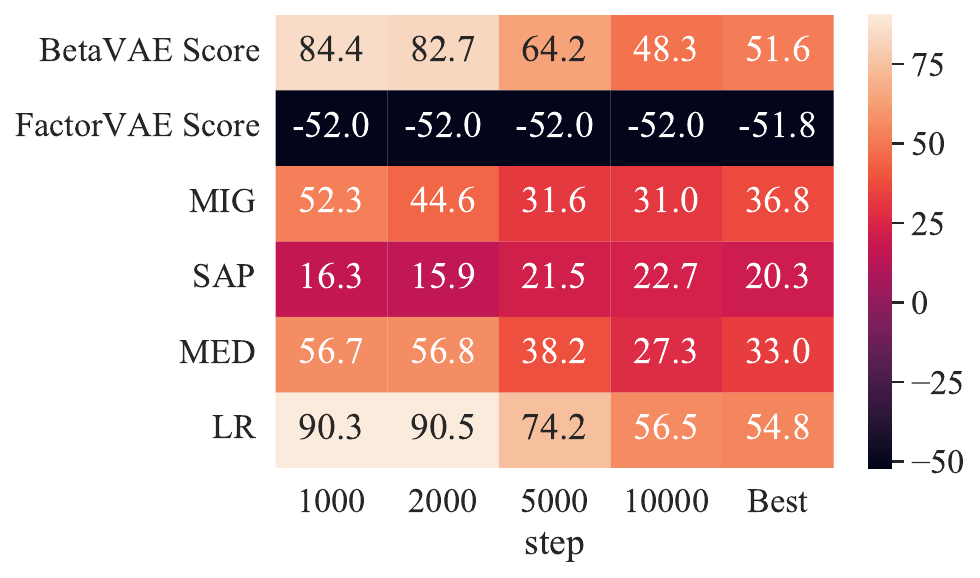}
    \caption{Correlations of SimSiam-WReN on \textit{3DShapes}.}
    \label{fig:3dshapes_simsiam_wren_heatmap}
\end{figure}
In the main text, we use BYOL as a representative of general-purpose representation learning methods. For completeness, here we introduce another general-purpose method, SimSiam \citep{chen2021exploring}. We modify the code of BYOL \footnote{\url{https://github.com/lucidrains/byol-pytorch.git}} to train SimSiams on \textit{3DShapes} with the following parameter grid: 
\begin{lstlisting}[language=Python]
[
  {'D': [512], 'lr': [3e-4, 3e-5], 
  'norm': [BatchNorm()], 'x_jit': [0.4, 0.8], 
  'p_gray': [0.3, 0.5, 0.7], 'x_crop': [0.6, 1.0]}
]
\end{lstlisting}
For each configuration, we run with 3 seeds. So finally, we yield 72 SimSiams. Then we use the same WReNs for DisVAEs and BYOLs as Stage-2 models. 

The results of SimSiam-WReN agree with our conclusions in the main text. As for the best performance, we have \awren{}=85.1\%, which is better than DisVAEs'. Figure~\ref{fig:3dshapes_simsiam_wren_heatmap} shows the correlations of downstream performance and representation properties. LR still correlates most for all considered steps. 

\begin{table*}[tb]
    \centering
    \begin{tabular}{c|l|r|r|r|r}
    \toprule
    Dataset & Stage1 &  \mwren{} & \awren{} & \mtrans{} & \atrans{} \\
\midrule
\multirow{2}{*}{\emph{3DShapes}} & DisVAEs & \makecell{AnnealedVAE\\@9600} & 
\makecell{\(\beta\)-VAE\\@8400(712)} &
\makecell{DIP-VAE-I\\@9900} & \makecell{\(\beta\)-TCVAE\\@9100(901)} \\
 & BYOL & \makecell{BYOL\\@10000} & \makecell{BYOL\\@8900(849)} & \makecell{BYOL\\@8600} &
\makecell{BYOL\\@8860(937)} \\
\midrule
\multirow{2}{*}{\emph{Abstract dSprites}} & DisVAEs & \makecell{\(\beta\)-VAE\\@9900} &
\makecell{DIP-VAE-I\\@8920(898)} & \makecell{DIP-VAE-I\\@9400} & \makecell{DIP-VAE-I\\@9380(172)} \\
 & BYOL & \makecell{BYOL\\@8100} & \makecell{BYOL\\@8320(1195)} &
\makecell{BYOL\\@8800} & \makecell{BYOL\\@8100(725)} \\
\bottomrule
    \end{tabular}
    \caption{{The model type and the step achieving the performance in Table \ref{tab:final_acc}. We report in the form of model@step. For \awren{} and \atrans{}, we report the mean steps and STDs.}}
    \label{tab:model_step_tab1}
\end{table*} 
\subsection{Additional Results of Final Performance}
\label{sec:additional_table1}
We have included the final performance of DisVAEs and BYOLs in Table \ref{tab:final_acc}. Additionally, we would like to provide further information on the specific DisVAEs and their corresponding steps that achieved the reported performance in Table \ref{tab:model_step_tab1}. It is evident that the optimal DisVAEs differ depending on the dataset and Stage-2 models used. While BYOL outperforms DisVAEs in terms of the earliest best performance in all cases except for \textit{3DShapes}-WReN, suggesting a faster learning process.

\subsection{Additional Results of Correlations}
\label{sec:additional_results_of_corr}
\begin{table*}[!t]
    \centering
    \resizebox{\textwidth}{!}
    {\begin{tabular}{c|l|r|r|r|r|r|r}
    \toprule
Dataset & Stage1 & BetaVAE & FactorVAE & MIG & SAP & MED & LR \\
\midrule
\multirow{2}{*}{\emph{3DShapes}} & DisVAEs & 93.7(7.7) & 82.3(11.1) &  25.5(15.1) &
6.5(3.8) & 45.4 (14.3) & 78.0(9.6) \\
 & BYOL & 99.9(0.3) & 96.1 (4.6) & 8.1(5.3) & 1.2(0.9) & 23.7 (7.4)&
96.6(1.8) \\
\midrule
\multirow{2}{*}{\makecell{\textit{Abstract} \\ \textit{dSprites}}} & DisVAEs & 62.3(14.1) & 49.1(10.5) & 13.3(7.0)
& 6.8(3.4) &15.5 (7.7) & 36.8(4.4) \\
 & BYOL & 63.6(17.0) & 62.4(11.8) & 2.6(1.8) &
0.5(0.3) &7.85 (2.9) & 43.0(8.2) \\
\bottomrule
    \end{tabular}}
    \caption{{Mean metric scores with STDs  of different Stage-1 models.}}
    \label{tab:mean_metric}
\end{table*}
In this part, we report additional results related to the correlations between representation metrics and downstream performance.

\begin{figure*}[!t]
    \begin{subfigure}{\textwidth}
        \centering
        \includegraphics[width=\textwidth]{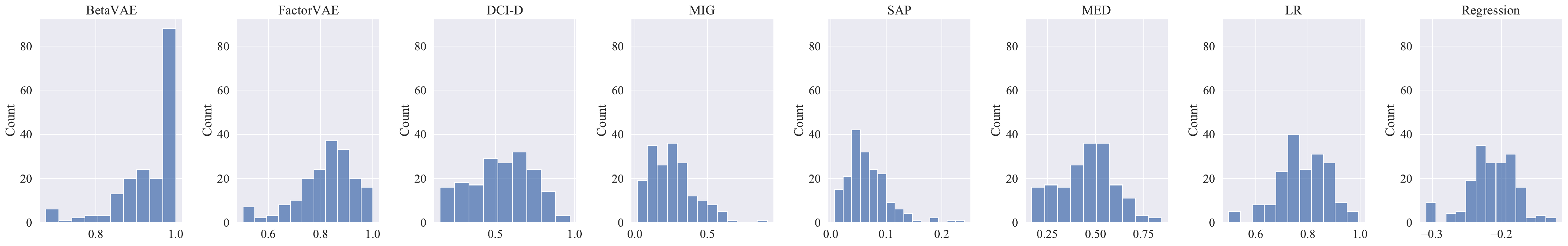}
        \caption{DisVAEs on \textit{3DShapes}}
        \label{fig:3dshpaes_vae_metric_hist}
    \end{subfigure} \\
    \begin{subfigure}{\textwidth}
        \centering
        \includegraphics[width=\textwidth]{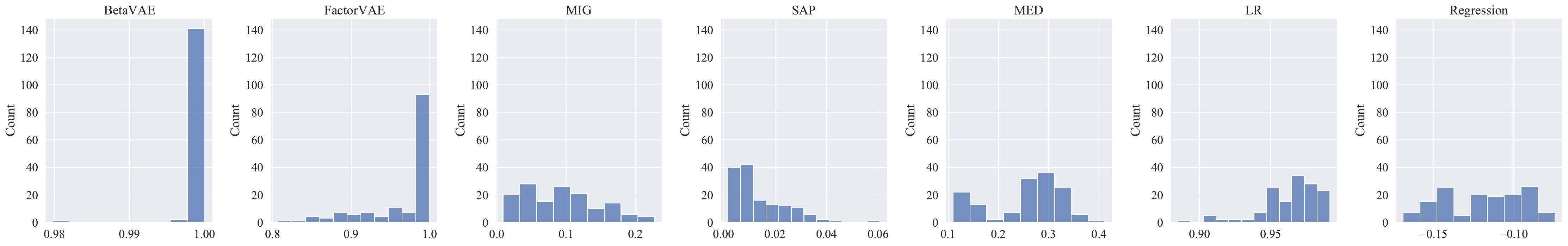}
        \caption{BYOL on \textit{3DShapes}}
        \label{fig:3dshpaes_byol_metric_hist}
    \end{subfigure} \\
    \begin{subfigure}{\textwidth}
        \centering
        \includegraphics[width=\textwidth]{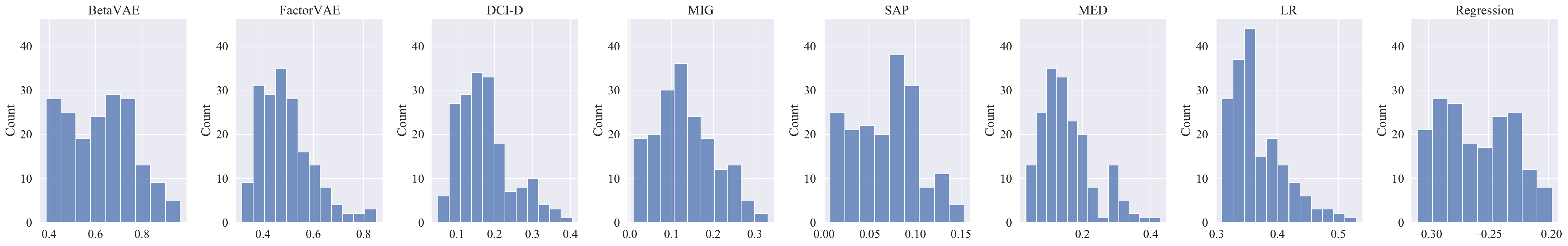}
        \caption{DisVAEs on \textit{Abstract dSprites}}
        \label{fig:dsprites_vae_metric_hist}
    \end{subfigure} \\
    \begin{subfigure}{\textwidth}
        \centering
        \includegraphics[width=\textwidth]{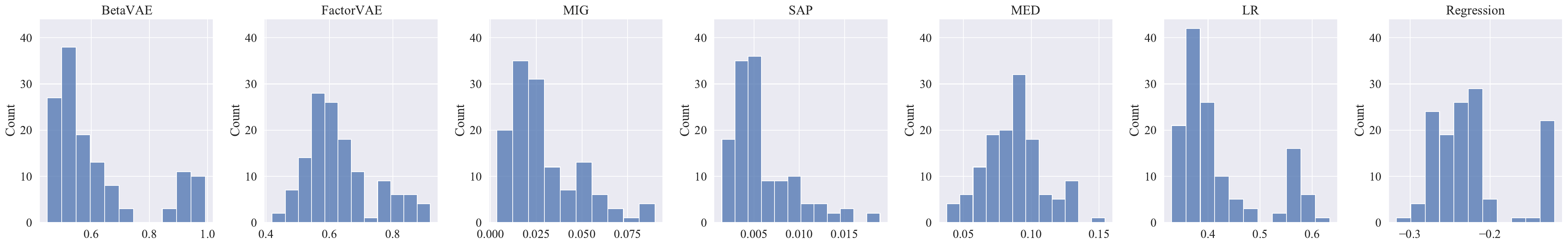}
        \caption{BYOL on \textit{Abstract dSprites}}
        \label{fig:dsprites_byol_metric_hist}
    \end{subfigure} 
    \caption{Histograms of metric scores of DisVAEs and BYOL on \textit{3dShapes} and \textit{Abstract dSprites}.}
    \label{fig:metric_hist}
\end{figure*}

\begin{figure*}[!th]
    \begin{subfigure}{\textwidth}
        \centering
        \includegraphics[width=\textwidth]{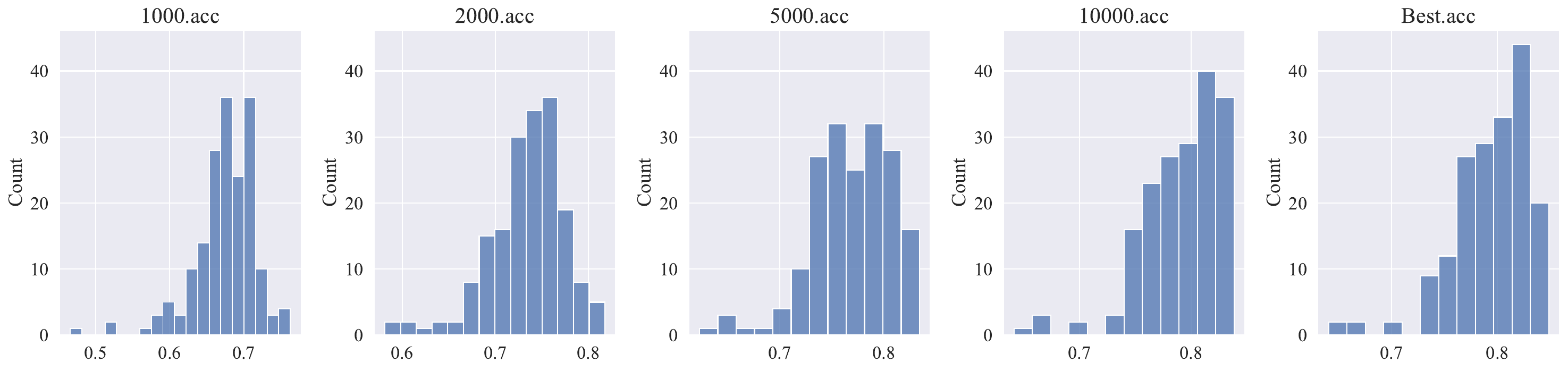}
        \caption{DisVAEs-WReN}
        \label{fig:3dshapes_vae_wren_acc_hist}
    \end{subfigure} \\
    \begin{subfigure}{\textwidth}
        \centering
        \includegraphics[width=\textwidth]{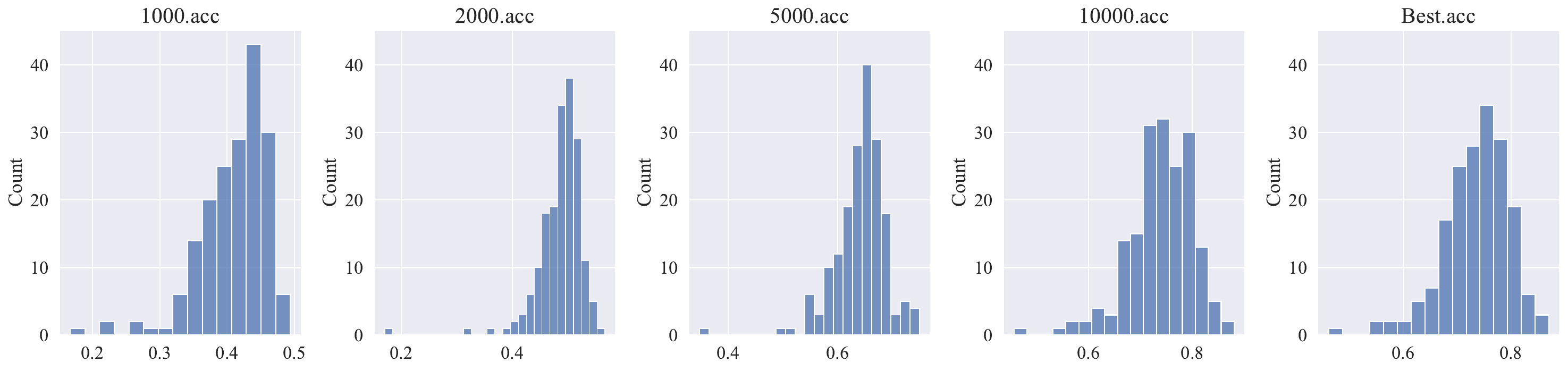}
        \caption{DisVAEs-Transformer}
        \label{fig:3dshapes_vae_trans_hist}
    \end{subfigure} \\
    \begin{subfigure}{\textwidth}
        \centering
        \includegraphics[width=\textwidth]{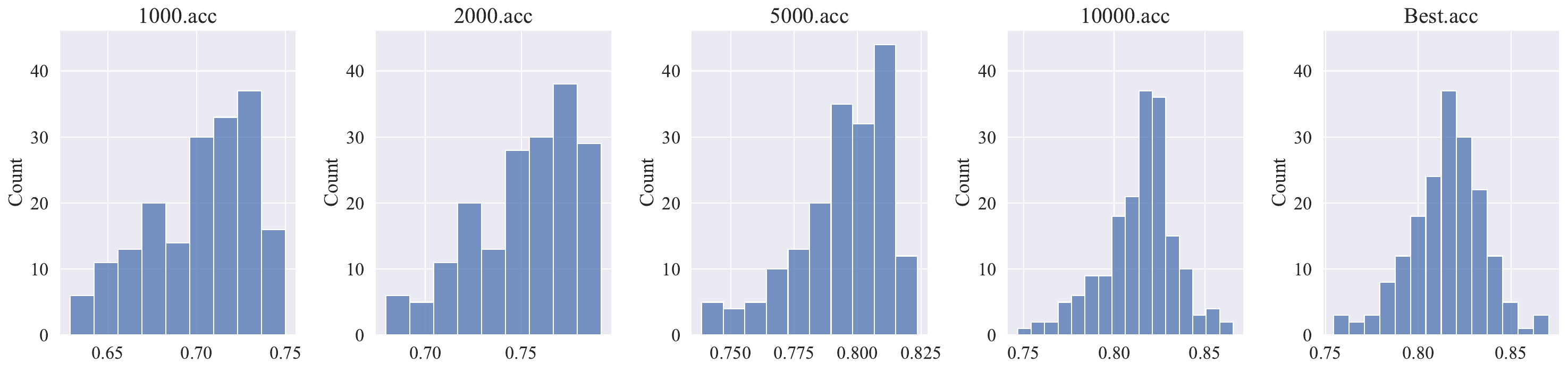}
        \caption{BYOL-WReN}
        \label{fig:3dshapes_byol_wren_acc_hist}
    \end{subfigure} \\
    \begin{subfigure}{\textwidth}
        \centering
        \includegraphics[width=\textwidth]{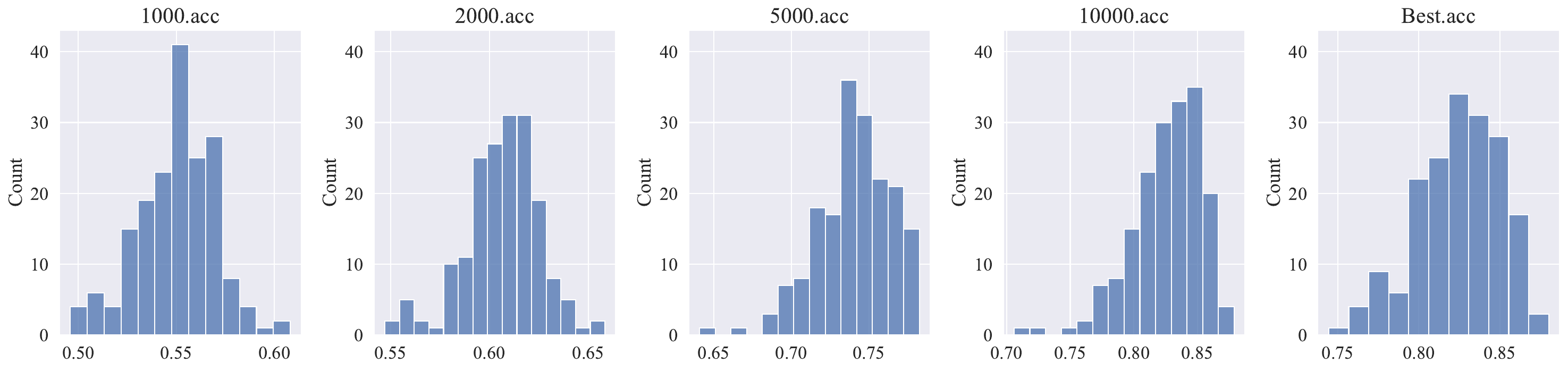}
        \caption{BYOL-Transformer}
        \label{fig:3dshapes_byol_trans_acc_hist}
    \end{subfigure} 
    \caption{Histograms of downstream accuracy for multiple steps on \textit{3DShapes}.}
    \label{fig:3dshapes_acc_hist}
\end{figure*}

\begin{figure*}[!ht]
    \begin{subfigure}{\textwidth}
        \centering
        \includegraphics[width=\textwidth]{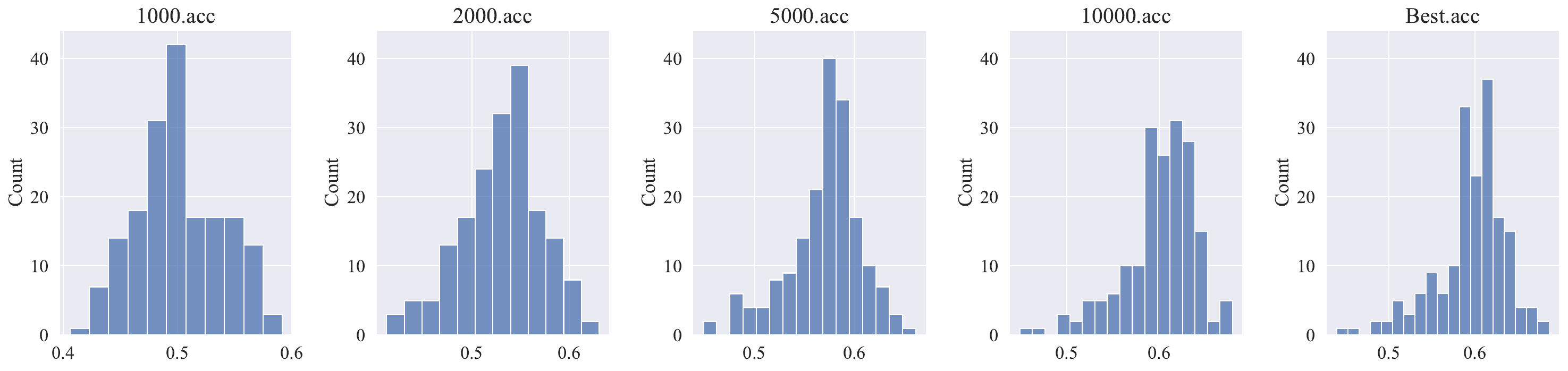}
        \caption{DisVAEs-WReN}
        \label{fig:dsprites_vae_wren_acc_hist}
    \end{subfigure} \\
    \begin{subfigure}{\textwidth}
        \centering
        \includegraphics[width=\textwidth]{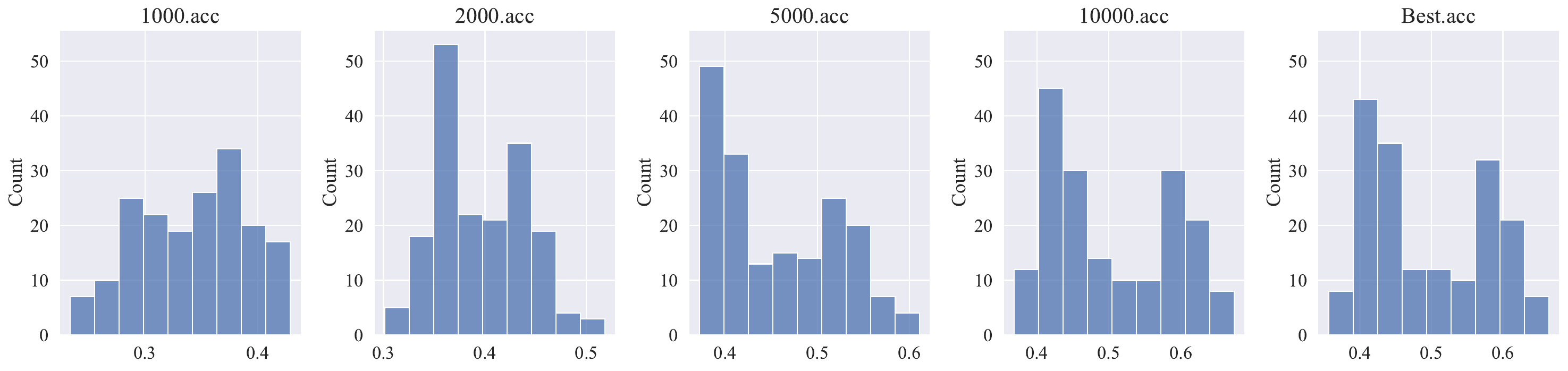}
        \caption{DisVAEs-Transformer}
        \label{fig:dsprites_vae_trans_hist}
    \end{subfigure} \\
    \begin{subfigure}{\textwidth}
        \centering
        \includegraphics[width=\textwidth]{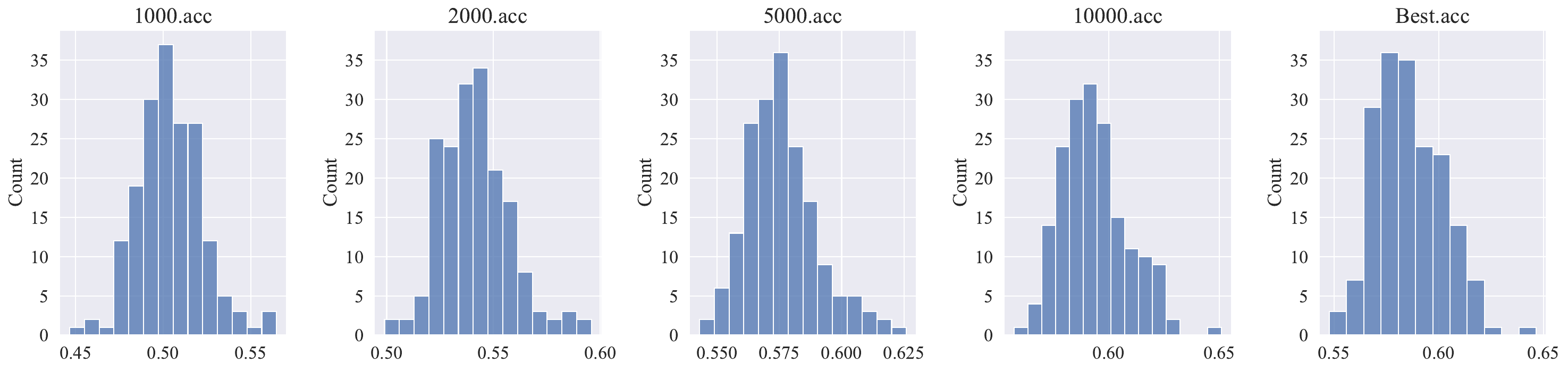}
        \caption{BYOL-WReN}
        \label{fig:dsprites_byol_wren_acc_hist}
    \end{subfigure} \\
    \begin{subfigure}{\textwidth}
        \centering
        \includegraphics[width=\textwidth]{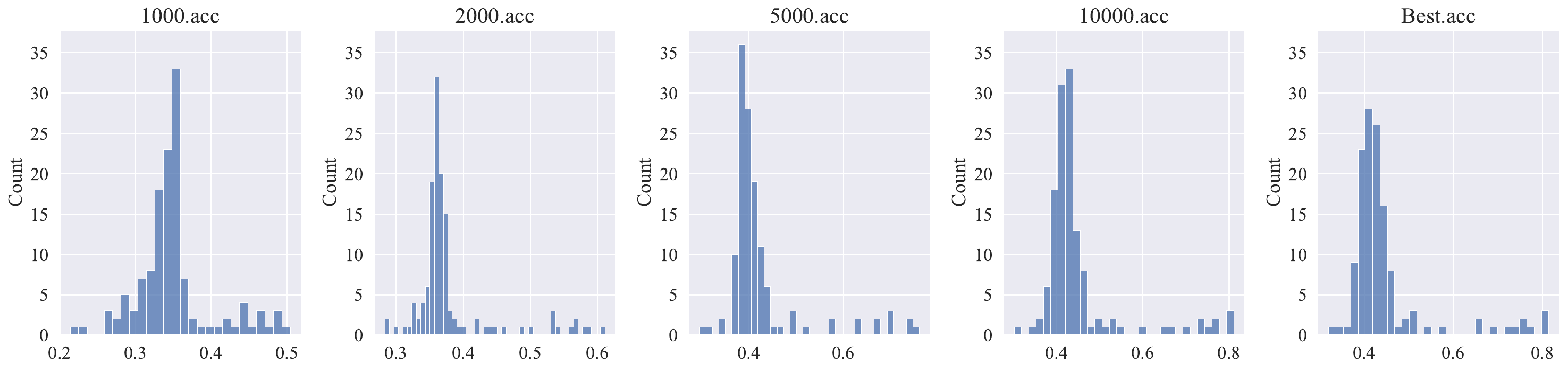}
        \caption{BYOL-Transformer}
        \label{fig:dsprites_byol_trans_acc_hist}
    \end{subfigure} 
    \caption{Histograms of downstream accuracy for multiple steps on \textit{Abstract dSprites}.}
    \label{fig:dsprites_acc_hist}
\end{figure*}

\textbf{Absolute values of metric scores and downstream accuracy.} We show the histograms as a sanity check of the distribution of metric scores and downstream accuracy. Figure~\ref{fig:metric_hist} presents the score distributions of each metric. We report the mean metric scores with STDs to depict the overall properties for Stage-1 models in Table~\ref{tab:mean_metric}. Figure~\ref{fig:3dshapes_acc_hist} and Figure~\ref{fig:dsprites_acc_hist} display the distributions of downstream performance. 

\textbf{Rank correlations.} This part contains additional results of rank correlations. On \textit{3DShapes}, Figure~\ref{fig:3dshapes_adj_heatmap} displays rank correlations between adjusted metrics and downstream accuracy, Figure~\ref{fig:3dshapes_self_heatmap} shows the overall correlation between metrics. On \textit{Abstract dSprites}, Figure~\ref{fig:dsprites_heatmap} shows correlations between metrics and downstream performance. Then Figure~\ref{fig:dsprites_adj_heatmap} presents correlations between adjusted metrics and downstream performance. Finally, Figure~\ref{fig:dsprites_self_heatmap} displays the overall correlations between metrics. 

\textbf{Plots of (metric score, downstream accuracy) pairs.}  Figures
\ref{fig:3dshapes_vae_wren_plot},~\ref{fig:3dshapes_vae_trans_plot},~\ref{fig:3dshapes_byol_wren_plot},~\ref{fig:3dshapes_byol_trans_plot},~\ref{fig:dsprites_vae_wren_plot},~\ref{fig:dsprites_vae_trans_plot},~\ref{fig:dsprites_byol_wren_plot}, and ~\ref{fig:dsprites_byol_trans_plot} 
provide an in-depth view of the correlations, where we plot $(\text{metrics}, \text{downstream accuracy})$ pairs.

\section{Limitations}
Our focus is on the essential downstream task of abstract visual reasoning, due to limited computation resources. We hope that our findings will be applicable to a broader range of downstream tasks despite this limitation.  We leave verification on other tasks for future work.

Furthermore, although we have embraced the broadly-recognized interpretation of disentanglement and its metrics, there exist more refined interpretations and assessments. Unfortunately, there is a shortage of corresponding techniques and measurements. We look forward to future developments to establish a more precise and structured definition of disentanglement that can be tested and validated in practical applications.

\newcommand{\happendix}{8.5em}

\begin{figure*}[!ht]
    \begin{subfigure}{.49\textwidth}
        \centering
        \includegraphics[height=\happendix{}]{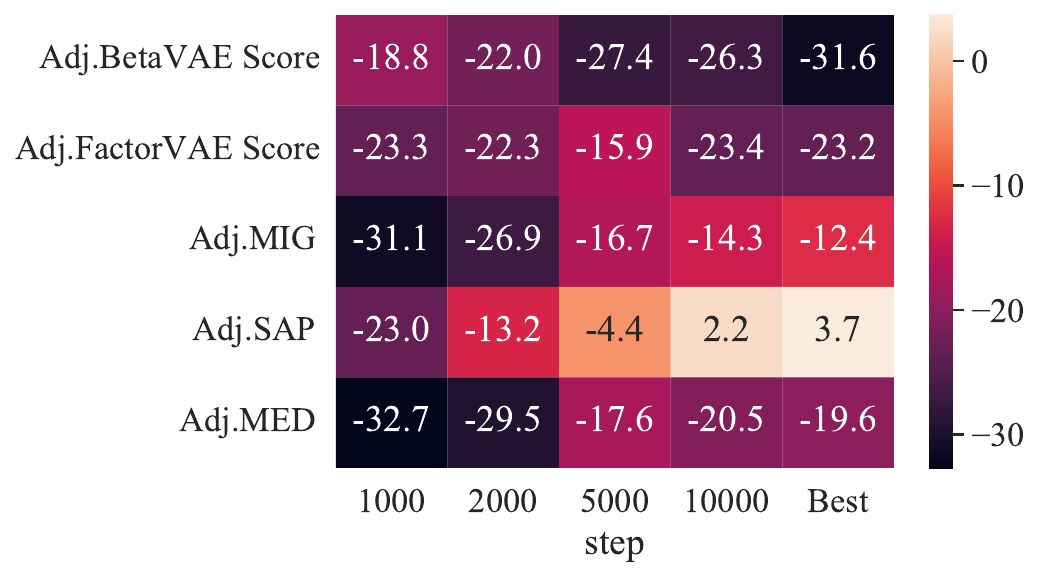}
        \caption{BYOL-WReN}
        \label{fig:3dshapes_WReN_byol_adj_heatmap}
    \end{subfigure} 
    \begin{subfigure}{.49\textwidth}
        \centering
        \includegraphics[height=\happendix{}]{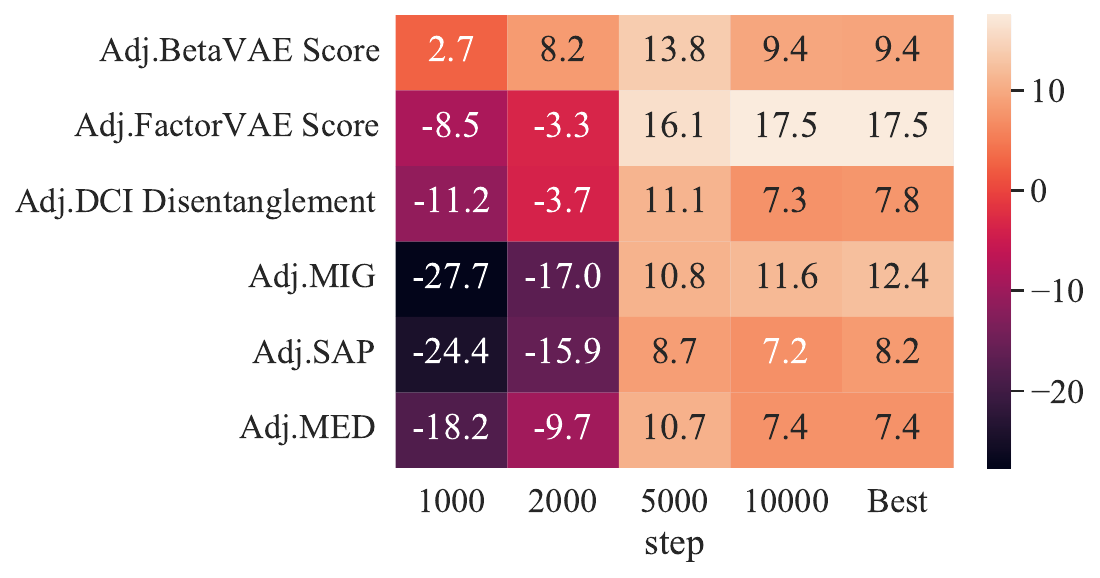}
        \caption{DisVAE-Transformer}
        \label{fig:3dshapes_transformer_vae_adj_heatmap}
    \end{subfigure}
    \\
    \begin{subfigure}{.49\textwidth}
        \centering
        \includegraphics[height=\happendix{}]{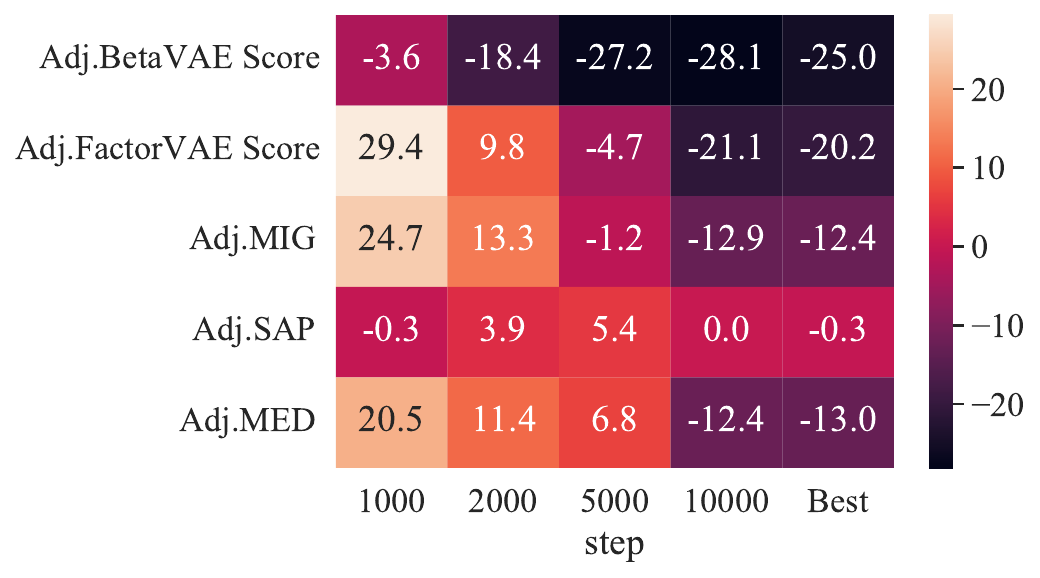}
        \caption{BYOL-Transformer}
        \label{fig:3dshapes_transformer_byol_adj_heatmap}
    \end{subfigure}
    
    \caption{Rank correlation between \awren{} or \atrans{} and adjusted metric scores on \textit{3DShapes}.}
    \label{fig:3dshapes_adj_heatmap}
\end{figure*}

\begin{figure*}[!ht]
    \begin{subfigure}{.49\textwidth}
        \centering
        \includegraphics[height=\happendix{}]{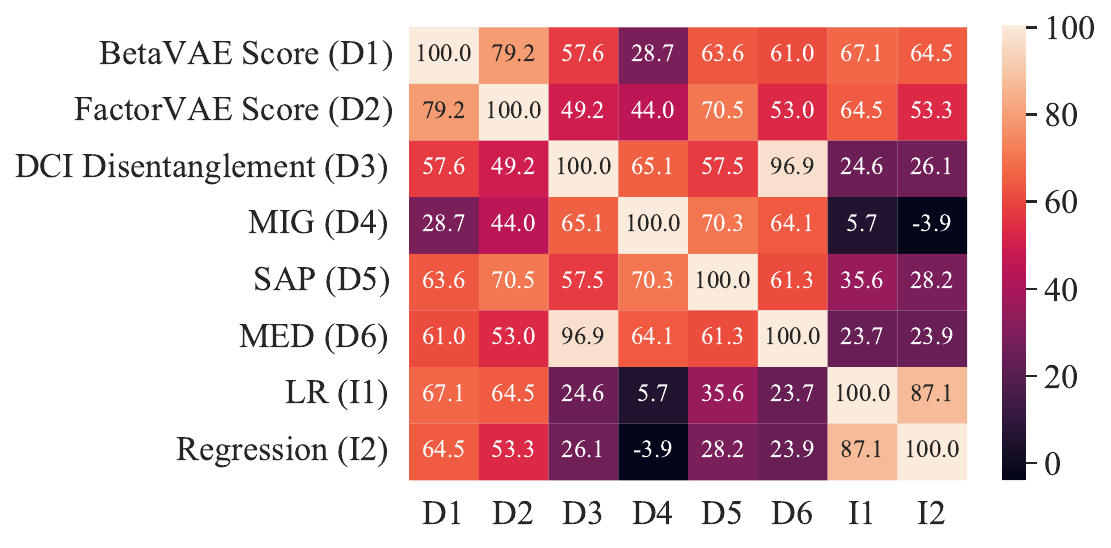}
        \caption{DisVAE}
        \label{fig:3dshapes_transformer_vae_self_heatmap}
    \end{subfigure}
    \begin{subfigure}{.49\textwidth}
        \centering
        \includegraphics[height=\happendix{}]{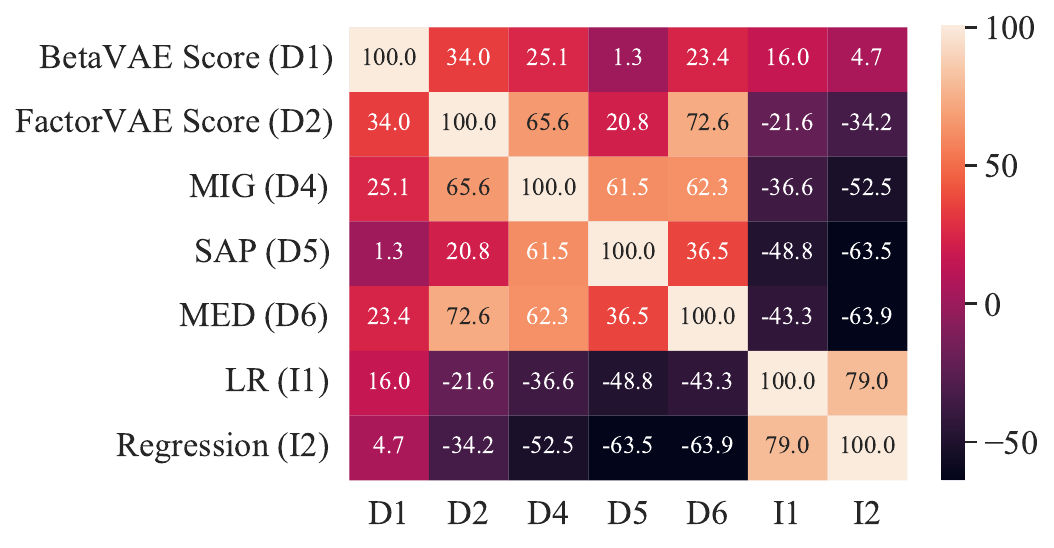}
        \caption{BYOL}
        \label{fig:3dshapes_WReN_byol_self_heatmap}
    \end{subfigure} 
    \caption{Overall correlation between metric scores on \textit{3DShapes}.}
    \label{fig:3dshapes_self_heatmap}
\end{figure*}

\begin{figure*}[!ht]
    \begin{subfigure}{.49\textwidth}
        \centering
        \includegraphics[height=\happendix{}]{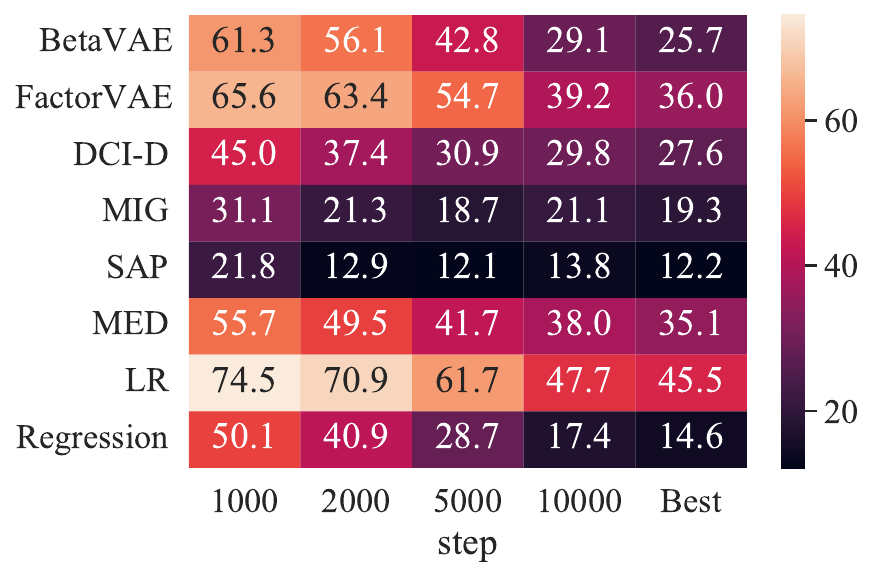}
        \caption{DisVAE-WReN}
        \label{fig:dsprites_WReN_vae_heatmap}
    \end{subfigure}
    \begin{subfigure}{.49\textwidth}
        \centering
        \includegraphics[height=\happendix{}]{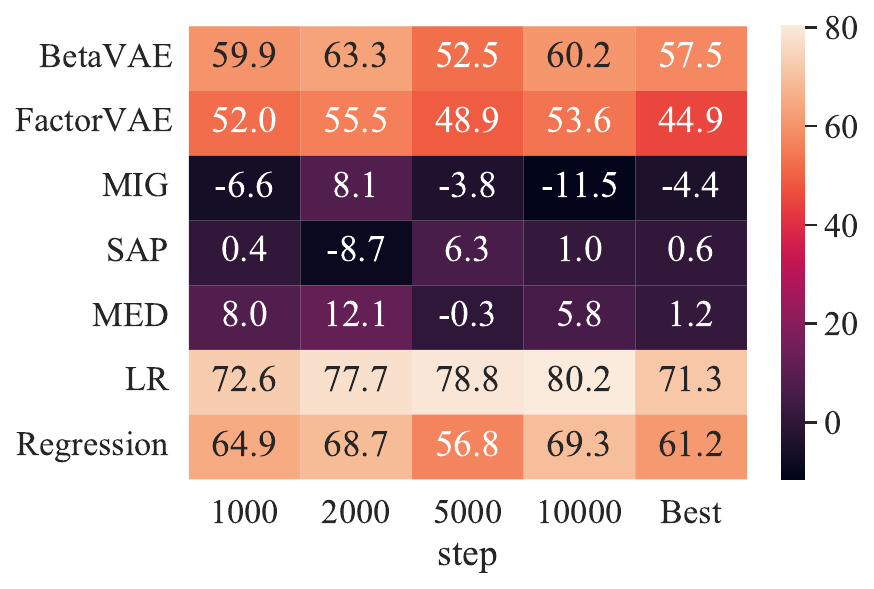}
        \caption{BYOL-WReN}
        \label{fig:dsprites_WReN_byol_heatmap}
    \end{subfigure} 
    \\
    \begin{subfigure}{.49\textwidth}
        \centering
        \includegraphics[height=\happendix{}]{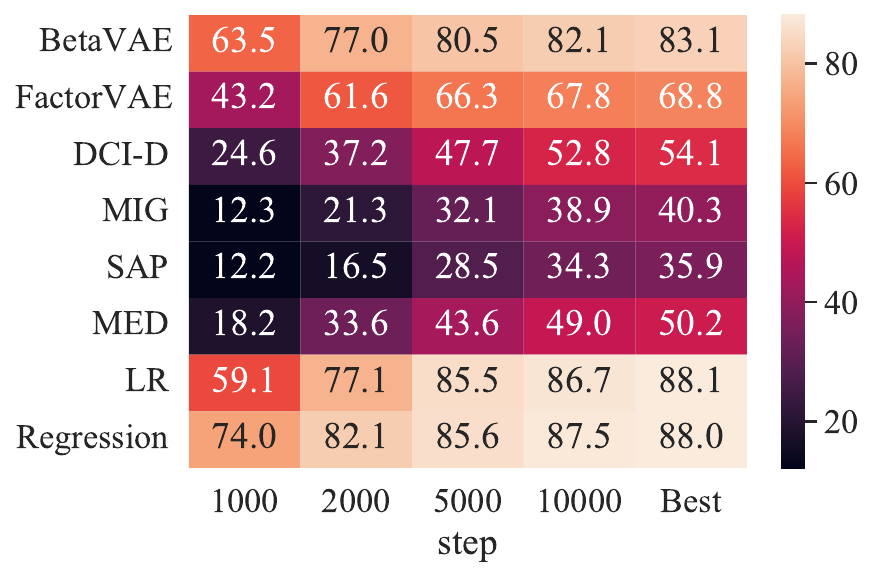}
        \caption{DisVAE-Transformer}
        \label{fig:dsprites_transformer_vae_heatmap}
    \end{subfigure}
    \begin{subfigure}{.49\textwidth}
        \centering
        \includegraphics[height=\happendix{}]{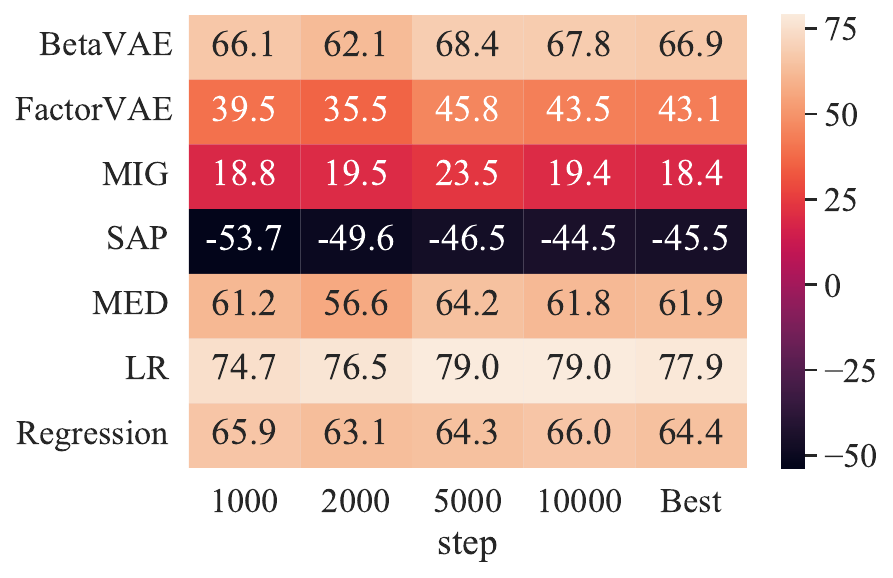}
        \caption{BYOL-Transformer}
        \label{fig:dsprites_transformer_byol_heatmap}
    \end{subfigure}
    
    \caption{Rank correlation between \awren{} or \atrans{} and metric scores on \textit{Abstract dSprites}.}
    \label{fig:dsprites_heatmap}
\end{figure*}

\begin{figure*}[!ht]
    \begin{subfigure}{.49\textwidth}
        \centering
        \includegraphics[height=\happendix{}]{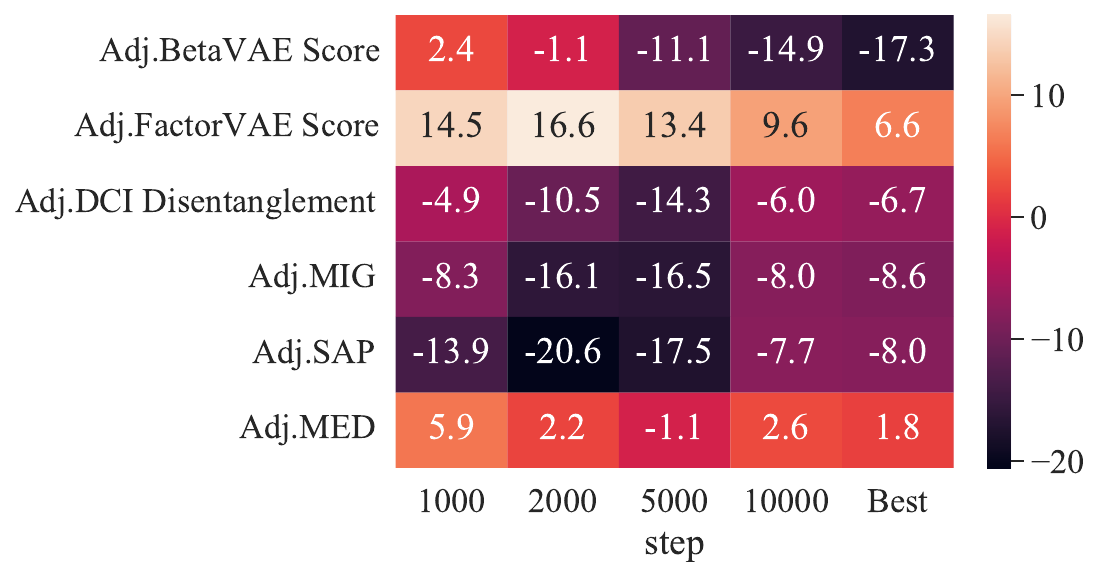}
        \caption{DisVAE-WReN}
        \label{fig:dsprites_WReN_vae_adj_heatmap}
    \end{subfigure}
    \begin{subfigure}{.49\textwidth}
        \centering
        \includegraphics[height=\happendix{}]{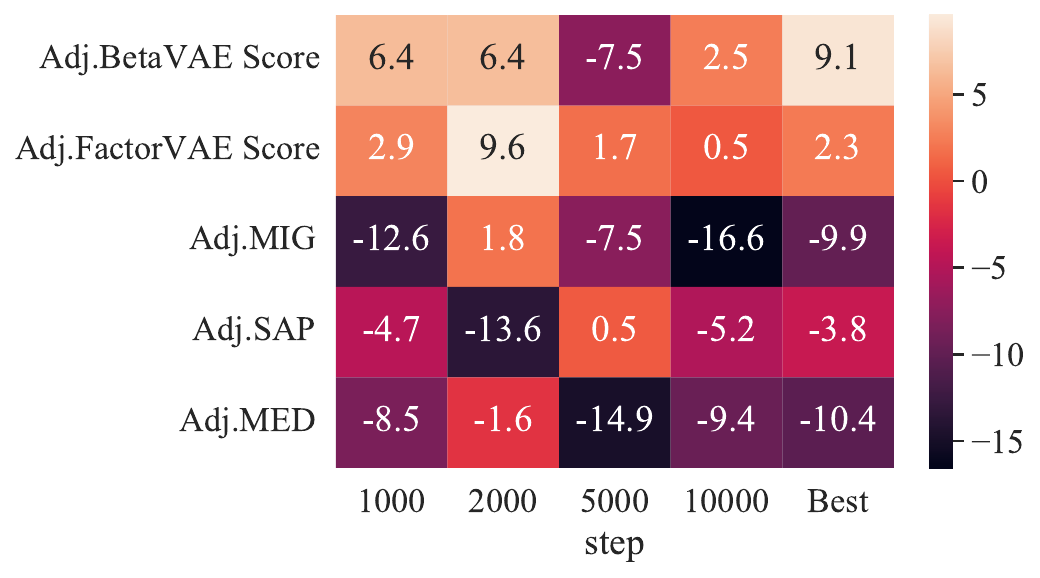}
        \caption{BYOL-WReN}
        \label{fig:dsprites_WReN_byol_adj_heatmap}
    \end{subfigure} 
    \\
    \begin{subfigure}{.49\textwidth}
        \centering
        \includegraphics[height=\happendix{}]{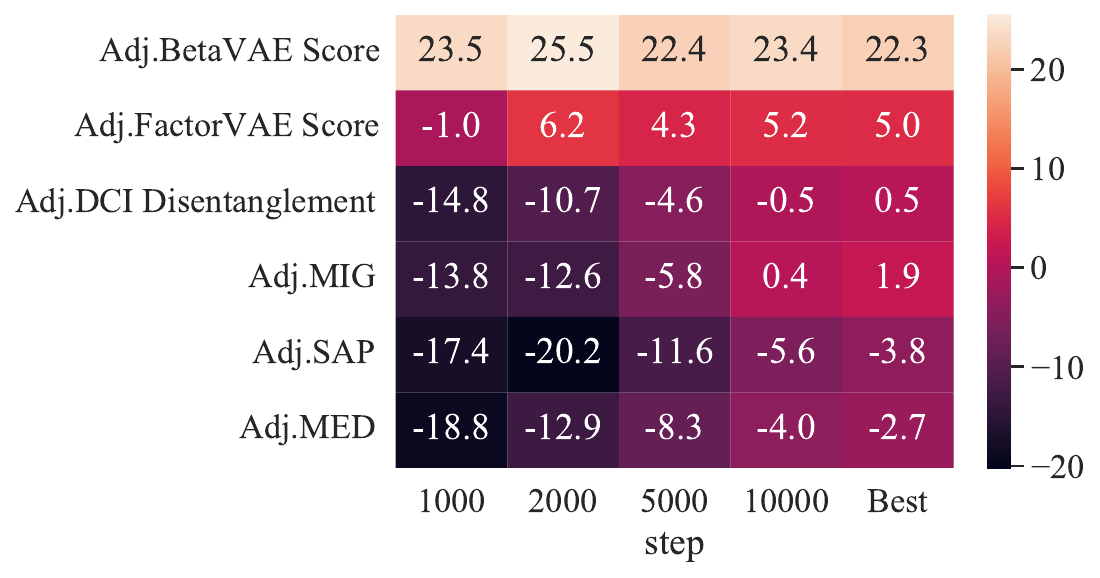}
        \caption{DisVAE-Transformer}
        \label{fig:dsprites_transformer_vae_adj_heatmap}
    \end{subfigure}
    \begin{subfigure}{.49\textwidth}
        \centering
        \includegraphics[height=\happendix{}]{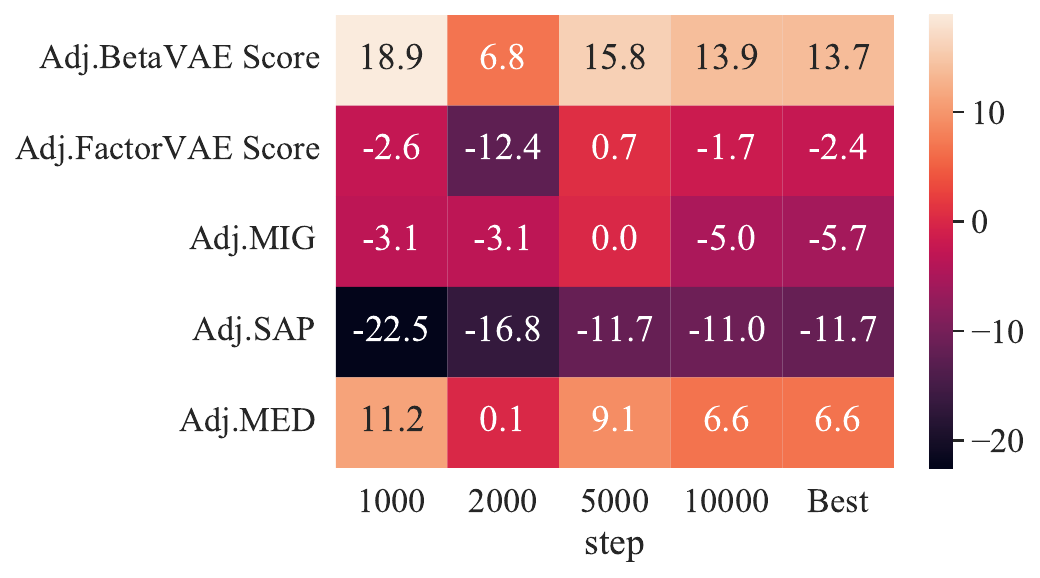}
        \caption{BYOL-Transformer}
        \label{fig:dsprites_transformer_byol_adj_heatmap}
    \end{subfigure}
    
    \caption{Rank correlation between \awren{} or \atrans{} and adjusted metric scores on \textit{Abstract dSprites}.}
    \label{fig:dsprites_adj_heatmap}
\end{figure*}

\begin{figure*}[!ht]
    \begin{subfigure}{.49\textwidth}
        \centering
        \includegraphics[height=\happendix{}]{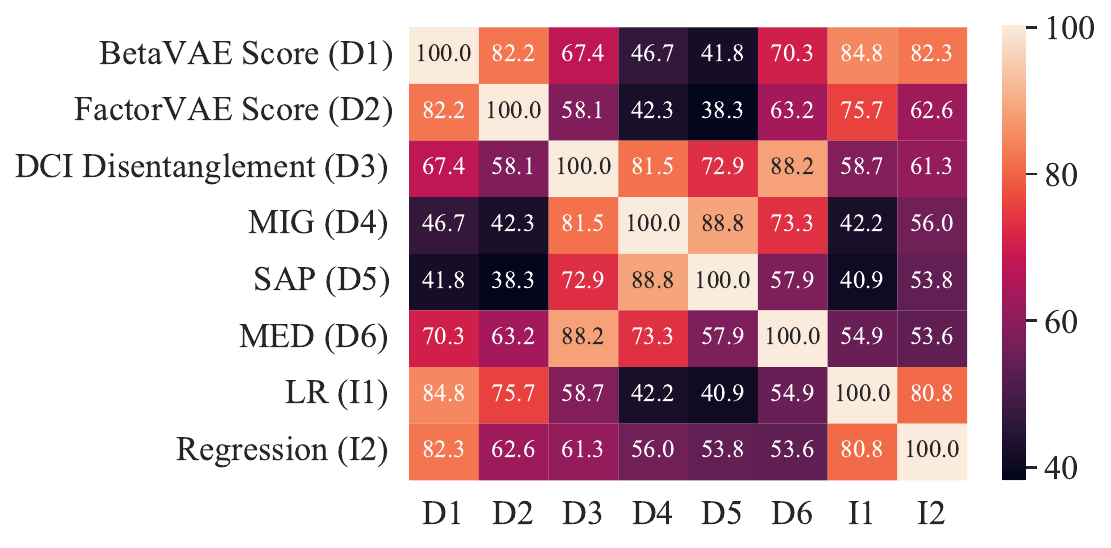}
        \caption{DisVAE}
        \label{fig:dsprites_transformer_vae_self_heatmap}
    \end{subfigure}
    \begin{subfigure}{.49\textwidth}
        \centering
        \includegraphics[height=\happendix{}]{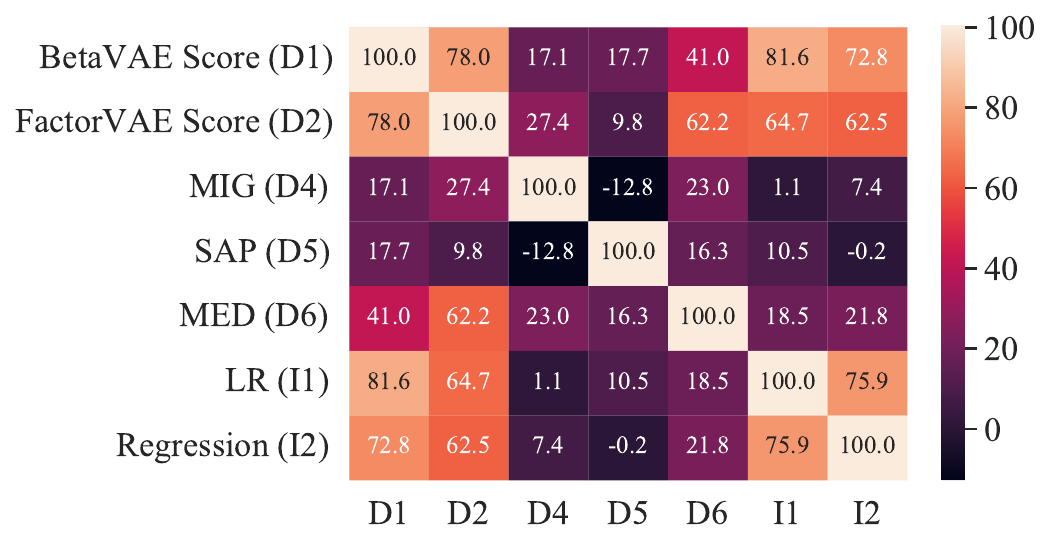}
        \caption{BYOL}
        \label{fig:dsprites_WReN_byol_self_heatmap}
    \end{subfigure} 
    \caption{Overall correlation between metric scores on \textit{Abstract dSprites}.}
    \label{fig:dsprites_self_heatmap}
\end{figure*}

\begin{figure*}[ht]
    \centering
    \includegraphics[width=\textwidth]{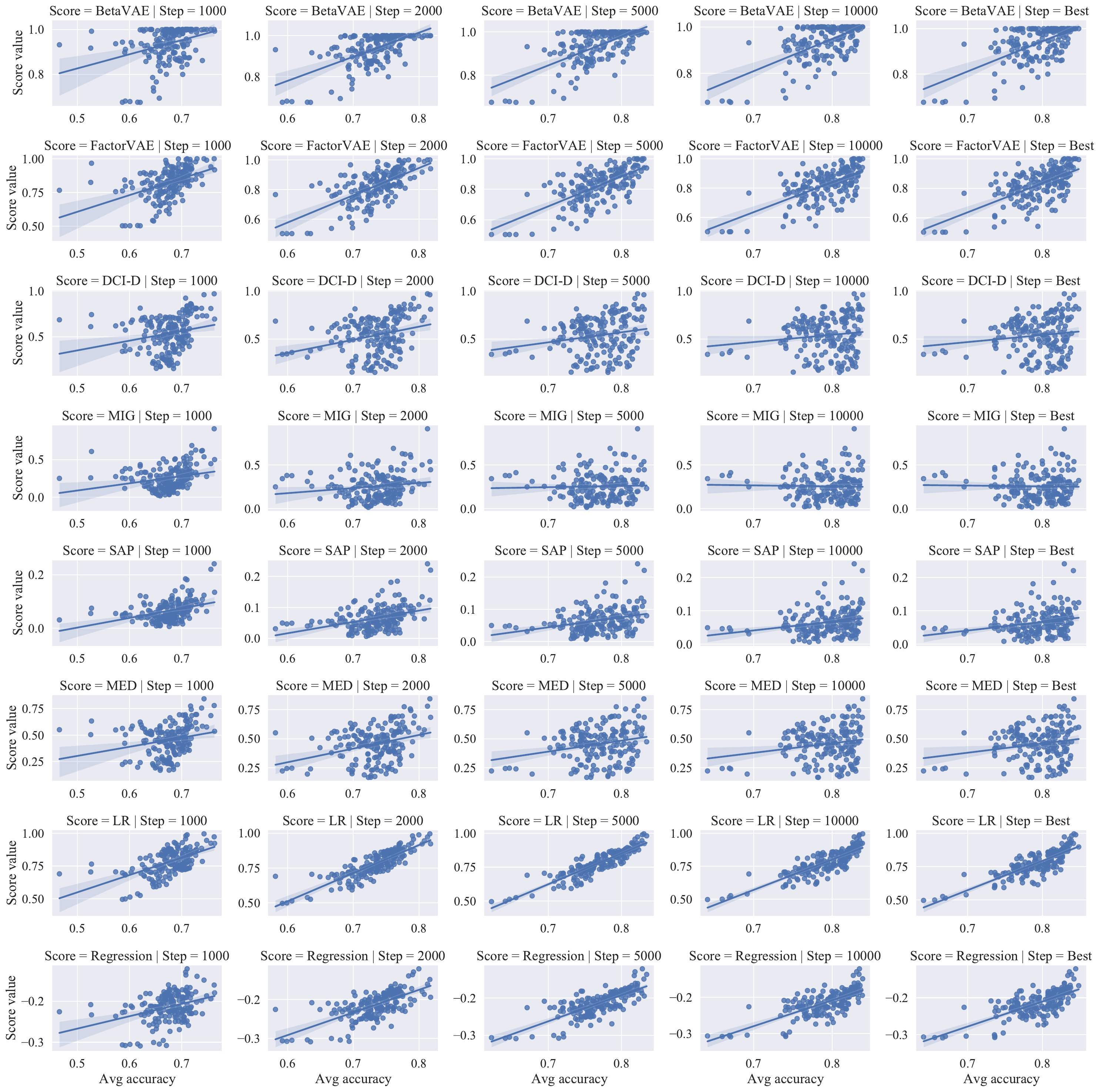}
    \caption{DisVAEs' metric scores v.s. \awren{} on \textit{3DShapes}.}
    \label{fig:3dshapes_vae_wren_plot}
\end{figure*}

\begin{figure*}[ht]
    \centering
    \includegraphics[width=\textwidth]{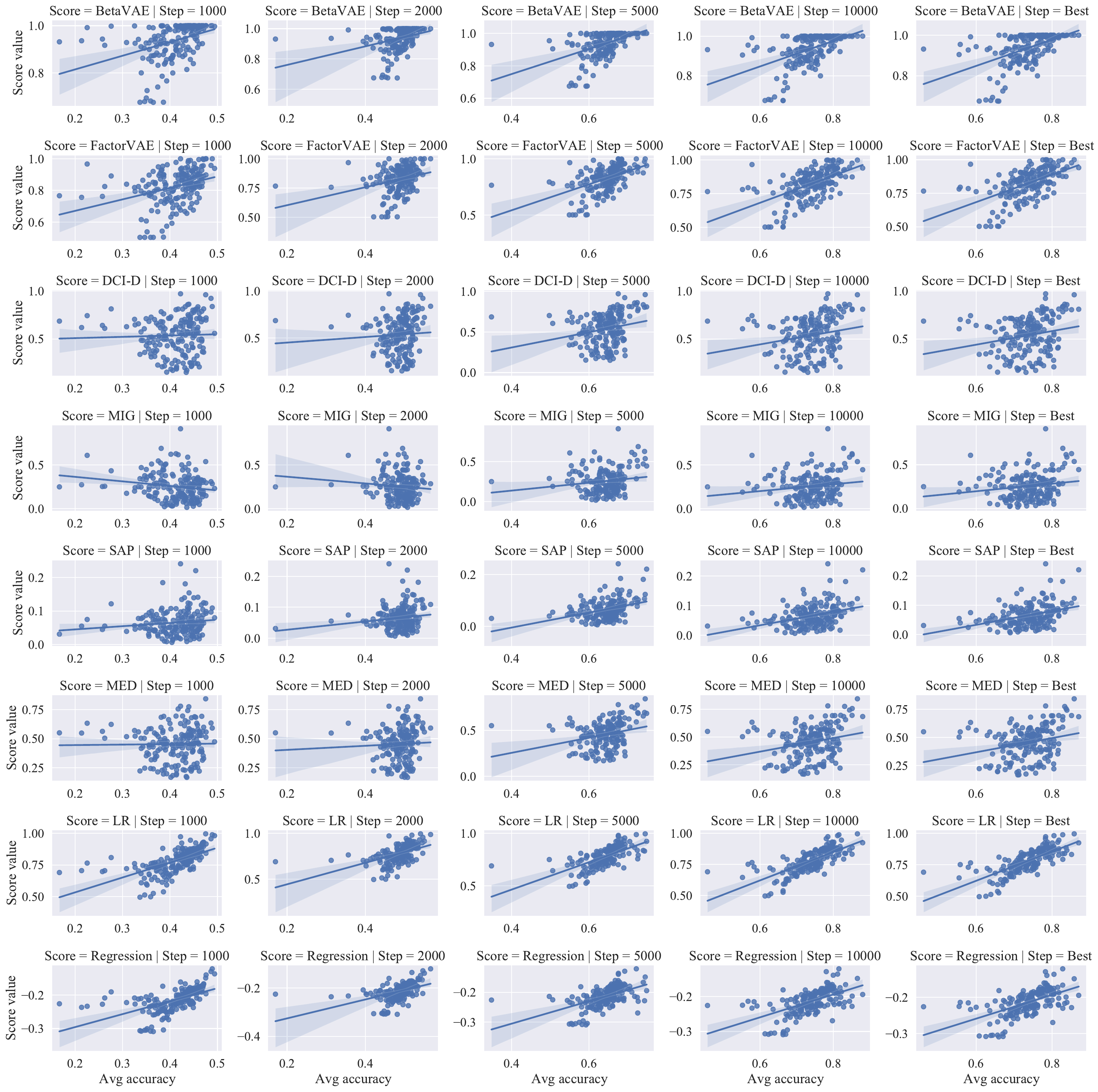}
    \caption{DisVAEs' metric scores v.s. \atrans{} on \textit{3DShapes}.}
    \label{fig:3dshapes_vae_trans_plot}
\end{figure*}

\begin{figure*}[ht]
    \centering
    \includegraphics[width=\textwidth]{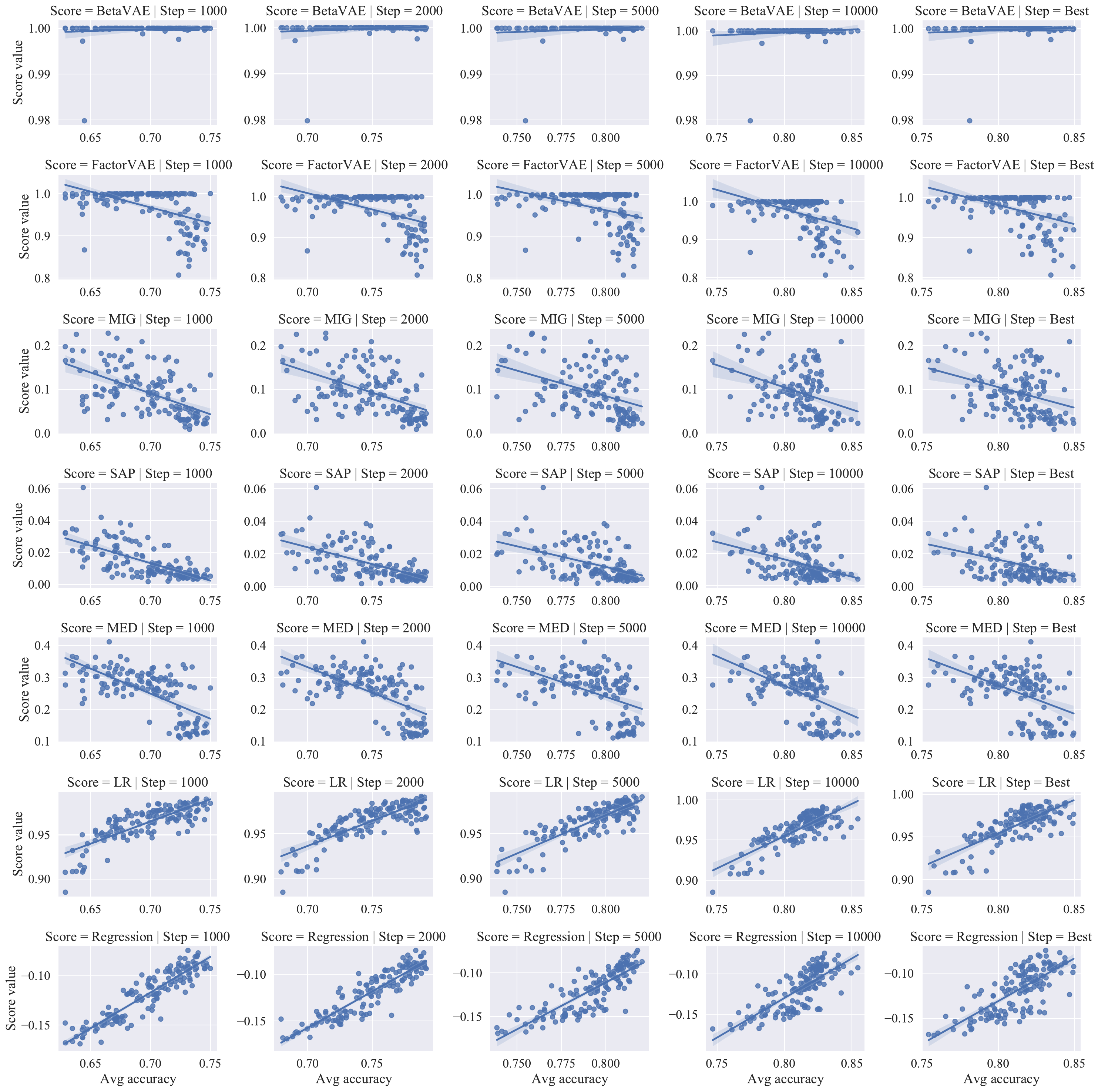}
    \caption{BYOLs' metric scores v.s. \awren{} on \textit{3DShapes}.}
    \label{fig:3dshapes_byol_wren_plot}
\end{figure*}

\begin{figure*}[ht]
    \centering
    \includegraphics[width=\textwidth]{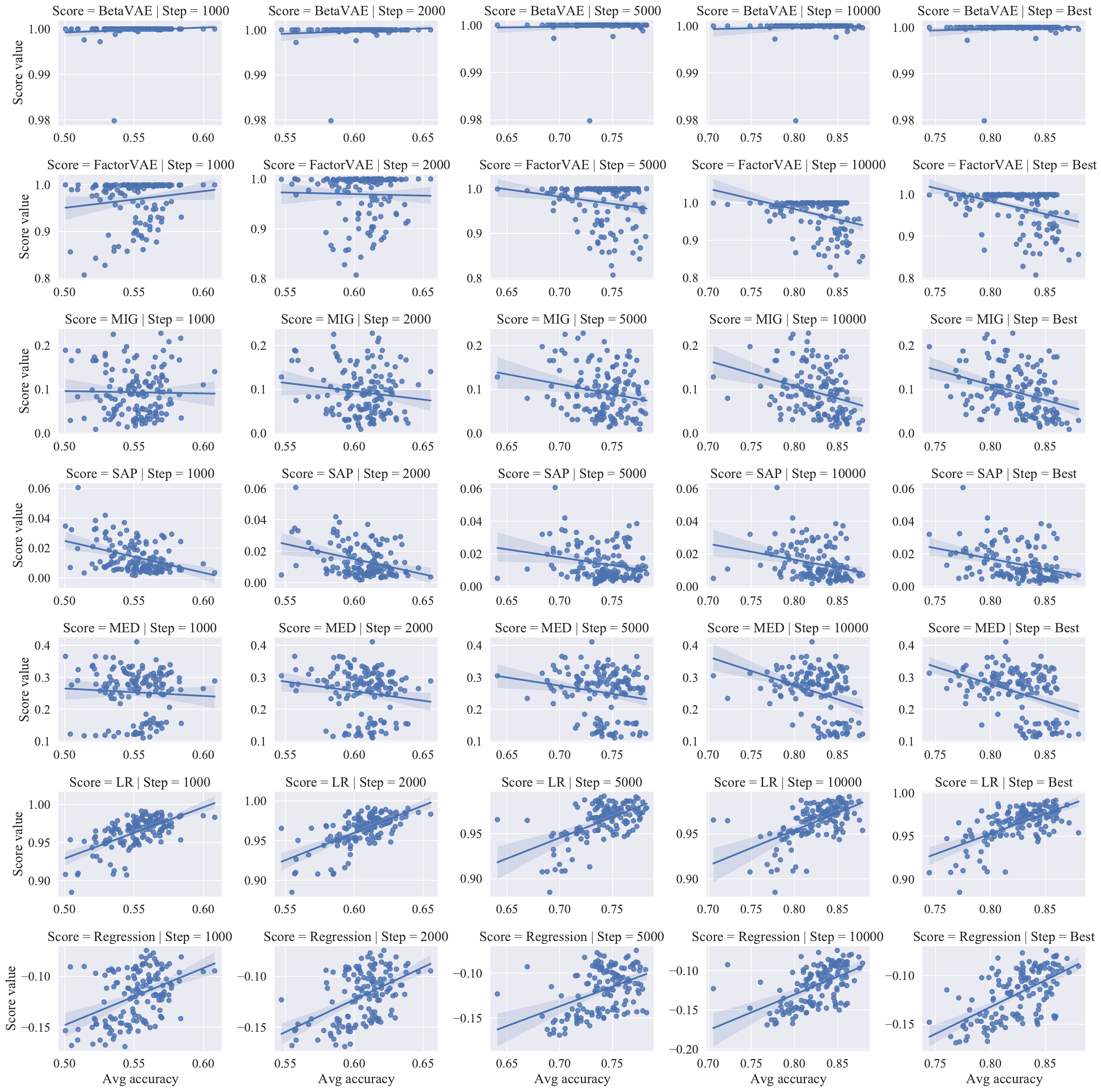}
    \caption{BYOLs' metric scores v.s. \atrans{} on \textit{3DShapes}.}
    \label{fig:3dshapes_byol_trans_plot}
\end{figure*}

\begin{figure*}[ht]
    \centering
    \includegraphics[width=\textwidth]{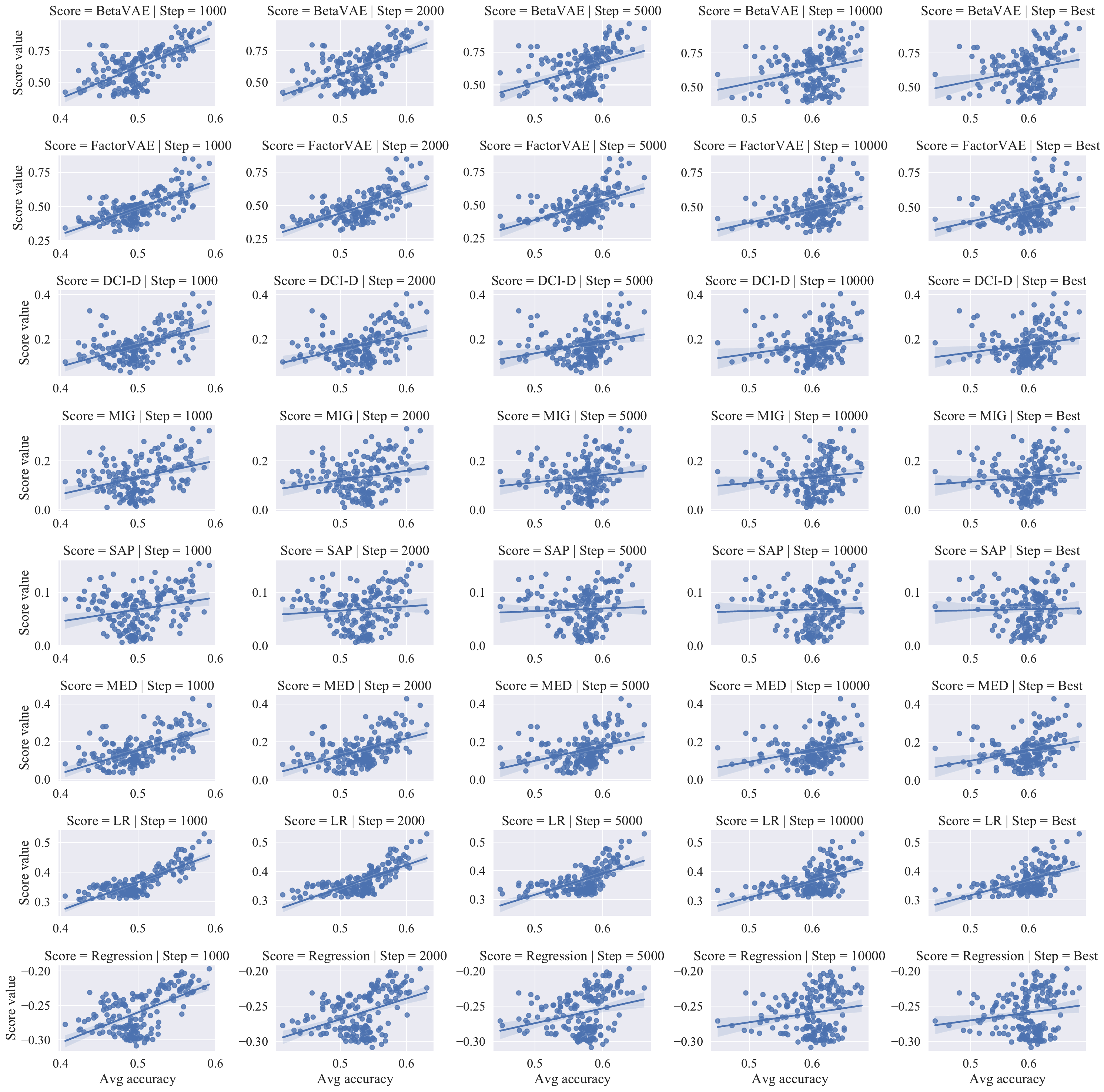}
    \caption{DisVAEs' metric scores v.s. \awren{} on \textit{Abstract dSprites}.}
    \label{fig:dsprites_vae_wren_plot}
\end{figure*}

\begin{figure*}[ht]
    \centering
    \includegraphics[width=\textwidth]{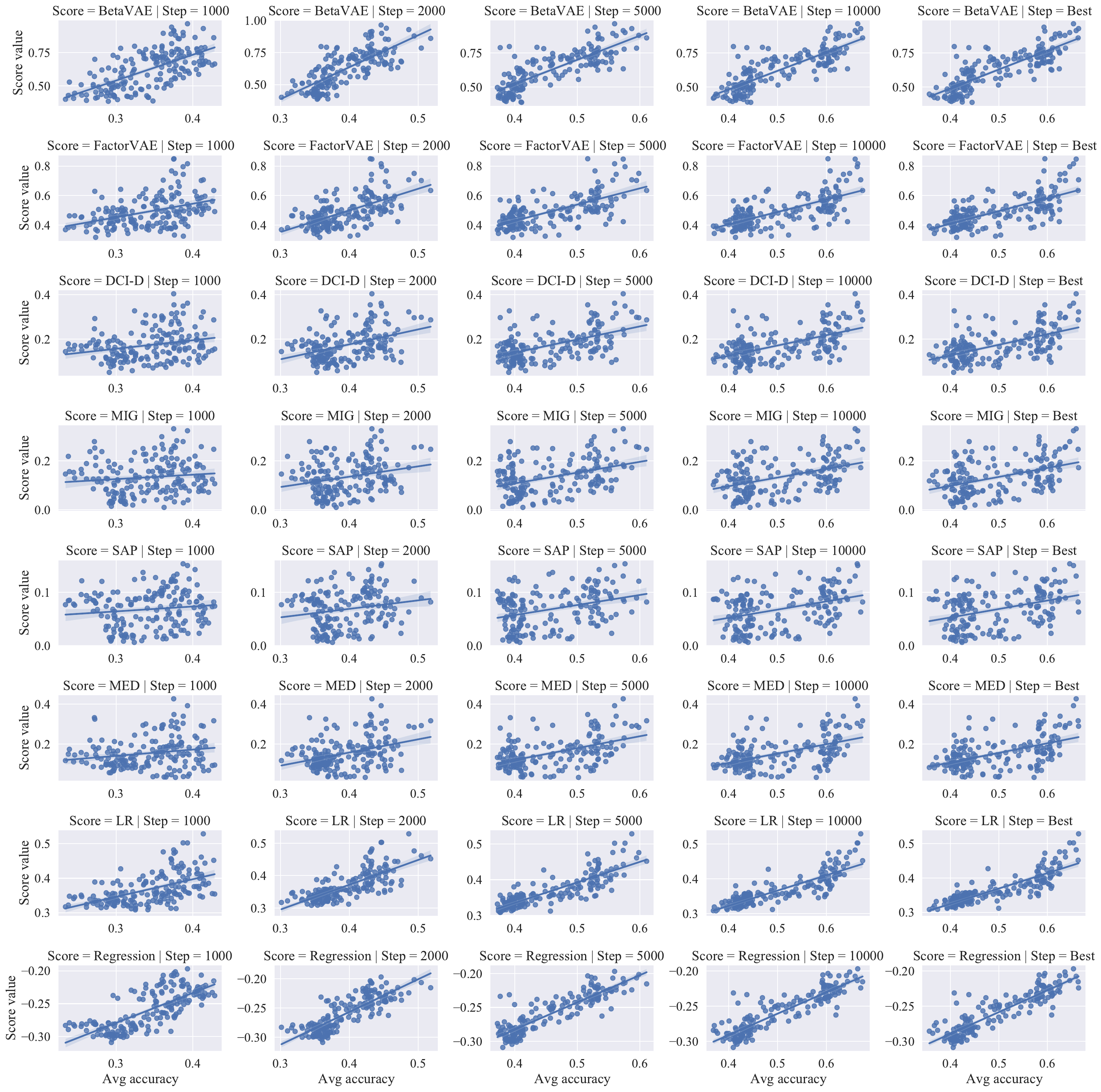}
    \caption{DisVAEs' metric scores v.s. \atrans{} on \textit{Abstract dSprites}.}
    \label{fig:dsprites_vae_trans_plot}
\end{figure*}

\begin{figure*}[ht]
    \centering
    \includegraphics[width=\textwidth]{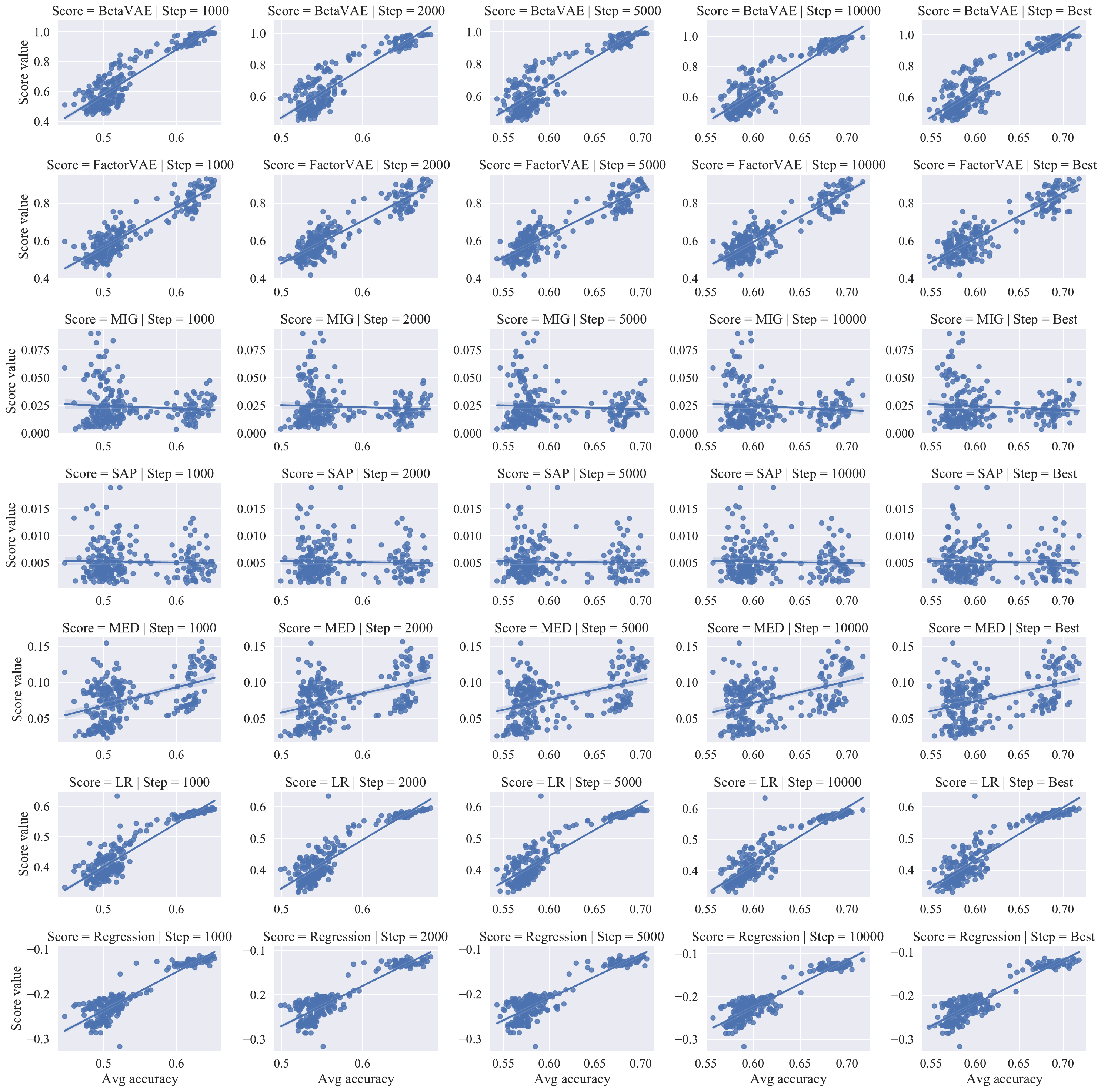}
    \caption{BYOLs' metric scores v.s. \awren{} on \textit{Abstract dSprites}.}
    \label{fig:dsprites_byol_wren_plot}
\end{figure*}

\begin{figure*}[ht]
    \centering
    \includegraphics[width=\textwidth]{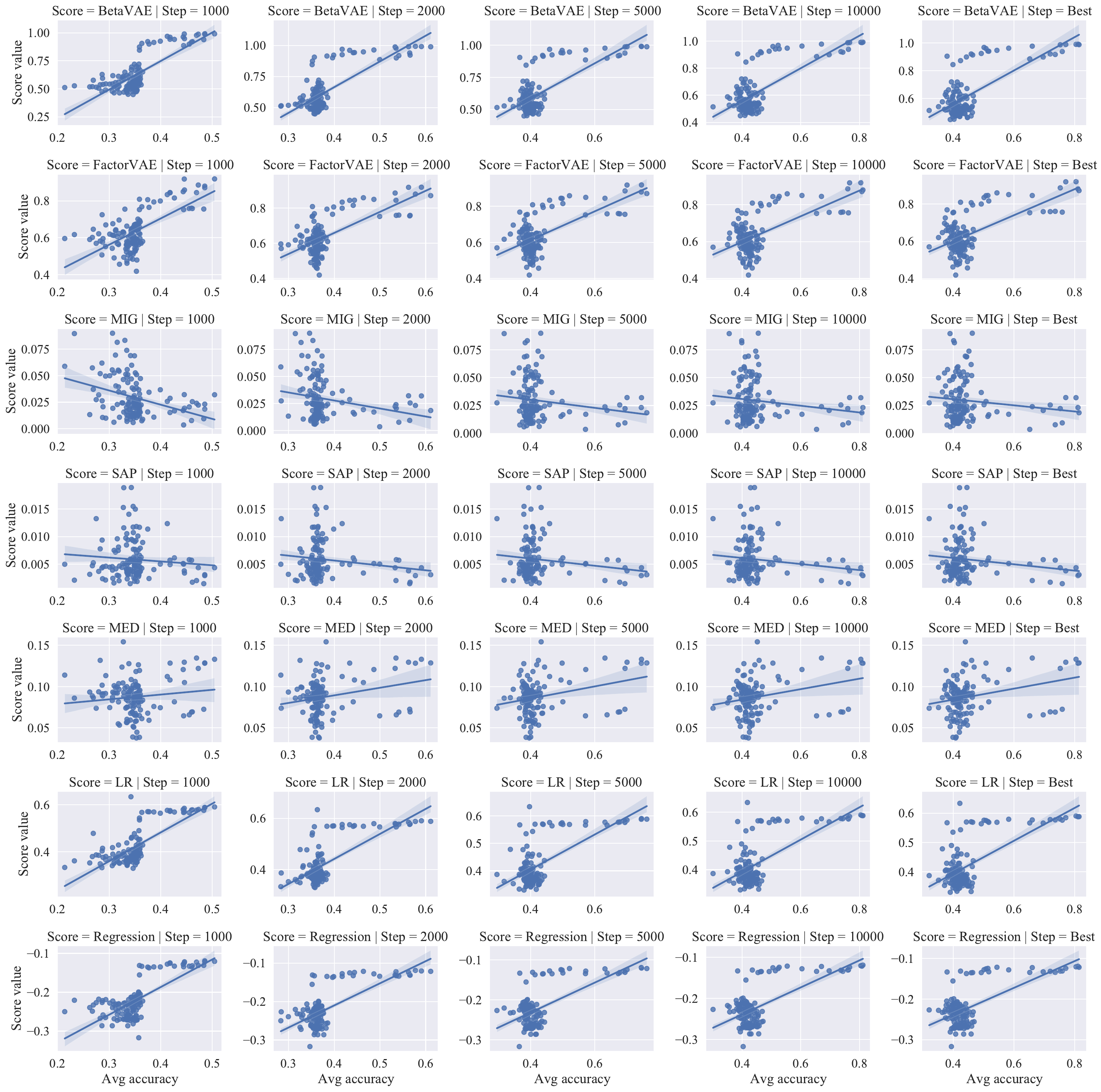}
    \caption{BYOLs' metric scores v.s. \atrans{} on \textit{Abstract dSprites}.}
    \label{fig:dsprites_byol_trans_plot}
\end{figure*}

\end{document}